%% file: main.tex
\documentclass[10pt,twocolumn,letterpaper]{article}

\usepackage{cvpr}              


\definecolor{cvprblue}{rgb}{0.21,0.49,0.74}
\usepackage[pagebackref,breaklinks,colorlinks,allcolors=cvprblue]{hyperref}

\usepackage{pifont}
\usepackage{booktabs}
\usepackage{float}
\usepackage{xcolor}
\usepackage{multirow}
\usepackage{subcaption}
\usepackage{soul}
\usepackage{hyperref}

\definecolor{darkgreen}{rgb}{0, 0.75, 0}  

\def\paperID{3249} 
\def\confName{CVPR}
\def\confYear{2025}

\title{OnlineAnySeg: Online Zero-Shot 3D Segmentation by Visual Foundation Model Guided 2D Mask Merging}

\author{
Yijie Tang\textsuperscript{1}$^{\ast}$ \thinspace Jiazhao Zhang\textsuperscript{3}$^{\ast}$ \thinspace Yuqing Lan\textsuperscript{1} \thinspace Yulan Guo\textsuperscript{4} \thinspace Dezun Dong\textsuperscript{1} \thinspace Chenyang Zhu\textsuperscript{1}$^{\dagger}$ \thinspace Kai Xu\textsuperscript{1,2}$^{\dagger}$ \\
\textsuperscript{1}National University of Defense Technology \quad
\textsuperscript{2}Xiangjiang Laboratory \\
\textsuperscript{3}CFCS, School of CS, Peking University \quad
\textsuperscript{4}Sun Yat-sen University
}

\begin{document}
\maketitle  

\renewcommand{\thefootnote}{}  
\footnote{$^{\ast}$Equal contribution. $^{\dagger}$Corresponding authors:}
\footnote{\emph{zhuchenyang07@nudt.edu.cn}, \thinspace \emph{kevin.kai.xu@gmail.com}}

\setlength{\abovecaptionskip}{3.5pt}  
\setlength{\belowcaptionskip}{1pt}  

\input{sec/0_abstract}
\input{sec/1_intro}
\input{sec/2_related_works}

\input{sec/3_method}

\input{sec/4_experiments}

\input{sec/5_conclusion}
\input{sec/6_acknowledgements}


\newpage

{
    \small
    \setlength{\bibsep}{0.0pt}  
    \setlength{\itemsep}{-0.5ex}  
    \bibliographystyle{ieeenat_fullname}
    \bibliography{main}
}

\end{document}




\setlength{\abovecaptionskip}{3.5pt}  
\setlength{\belowcaptionskip}{1pt}  

\input{sec/X_suppl_1_main}

\newpage

{
    \small
    \setlength{\bibsep}{0.0pt}  
    \setlength{\itemsep}{-0.5ex}  
    \bibliographystyle{ieeenat_fullname}
    \bibliography{main}
}

%% file: sec/0_abstract.tex
\begin{abstract}

\definecolor{url_color}{rgb}{0.96, 0.22 , 0.6}
\hypersetup{
    colorlinks=true,       
    linkcolor=url_color,        
    citecolor=url_color,        
    urlcolor=url_color          
}

 Online zero-shot 3D instance segmentation of a progressively reconstructed scene is both a critical and challenging task for embodied applications. With the success of visual foundation models (VFMs) in the image domain, leveraging 2D priors to address 3D online segmentation has become a prominent research focus. Since segmentation results provided by 2D priors often require spatial consistency to be lifted into final 3D segmentation, an efficient method for identifying spatial overlap among 2D masks is essential—yet existing methods rarely achieve this in real time, mainly limiting its use to offline approaches. To address this, we propose an efficient method that lifts 2D masks generated by VFMs into a unified 3D instance using a hashing technique. By employing voxel hashing for efficient 3D scene querying, our approach reduces the time complexity of costly spatial overlap queries from $O(n^2)$ to $O(n)$. Accurate spatial associations further enable 3D merging of 2D masks through simple similarity-based filtering in a zero-shot manner, making our approach more robust to incomplete and noisy data. Evaluated on the ScanNet200 and SceneNN benchmarks, our approach achieves state-of-the-art performance in online, zero-shot 3D instance segmentation with leading efficiency. The project page is at \href{https://yjtang249.github.io/OnlineAnySeg}{https://yjtang249.github.io/OnlineAnySeg}.

\end{abstract}

%% file: sec/1_intro.tex
\section{Introduction}
\label{sec:intro}

3D instance segmentation of an online reconstructed scene is a difficult yet important task for robotic scene exploration and understanding. In contrast to offline segmentation, online segmentation must deal with the incompleteness and ambiguity of an incrementally reconstructed scene under real-time constraints. With the availability of labeled 3D scene datasets such as ScanNet200~\cite{rozenberszki2022language}, existing methods have achieved accurate online segmentation through supervised training over a closed set of object categories~\cite{mccormac2017semanticfusion, zhang2020fusion, dai20183dmv, hou20193d, huang2021supervoxel}. The recent embodied applications, however, call for online 3D segmentation in the open-vocabulary setting, making the problem more challenging.

\input{fig/fig_teaser}

The recent advances in visual foundation models (VFMs), such as SAM~\cite{kirillov2023segment} and OpenSEED~\cite{zhang2023simple}, have demonstrated strong zero-shot ability in 2D image segmentation. Leveraging the 2D priors of VFMs to address open-vocabulary 3D instance segmentation is a promising direction. Specifically, one can merge the 2D instance masks of several viewpoints based on multi-view consistency to form a unified 3D segmentation result.
The two key challenges here are (1) how to find 3D spatial associations of 2D masks from different views and (2) how to determine merges of the associated masks. The former is computationally costly and has been the main bottleneck of real-time performance. 
A walkaround is to learn to predict the merging of all pairs of masks without explicitly finding mask associations~\cite{xu2024embodiedsam}. This approach, however, tends to exhibit a noticeable performance drop when handling incomplete data during the online, incremental reconstruction, because the model has been trained offline usually with complete scenes.
We, therefore, argue that an efficient organization of 2D masks is still essential for fast determination of their multi-view consistency.

In this work, we propose a simple yet effective strategy for the online organization and merging of instance masks (obtained by CropFormer~\cite{qi2023high}) based on a hashing technique.
To solve the multi-view association problem, we employ a hashed voxel volume for scene representation due to its good query efficiency, based on the VoxelHashing framework~\cite{niessner2013real}. In addition to storing TSDF values at each voxel, the hashing table also maintains IDs of 2D masks back-projected to the corresponding voxels. This allows us to fast query mask overlap. 
Mask merging updates the mask IDs in the relevant voxel entries of the hash table, which can be time-consuming due to the large number of voxels. To address this, we designed a mapping table for mask IDs. During merging, we simply update the mask mappings in this table to efficiently refresh the mask information.
Since the number of masks is significantly smaller than that of voxels, our method achieves high efficiency by avoiding the overhead of frequent operations on voxel hashes.

With the efficient maintenance of mask associations, the next step is to determine mask merging based on mask similarity.
In particular, mask similarity is measured based on mask overlap, semantic similarity, and geometric similarity. 
Mask overlap can be efficiently determined based on our hash-based scene representation. Semantic and geometric similarities can be computed based on the open-vocabulary features extracted by VFMs~\cite{qi2023high, radford2021learning} and point cloud correspondence features~\cite{choy2019fully}, respectively.

Through extensive evaluations on the ScanNet200 and SceneNN benchmarks, we demonstrate that our method achieves SOTA performance of online, zero-shot 3D instance segmentation. Our contributions include:

\begin{itemize}
    \item We propose an efficient data structure for organizing sequential 2D masks, which can incrementally maintain the spatial associations between all the masks in real-time.
    \item We design a zero-shot online mask merging strategy. By leveraging spatial overlap and multimodal similarity through collaborative filtering, our approach eliminates the dependency on training data, enabling it to maintain good performance even in incomplete scanned scenes.
    \item Our method performs comparably to offline methods~\cite{yan2024maskclustering} and gains notable improvements over the SOTA online method on the publicly available benchmark, running at 15 FPS.
\end{itemize}

%% file: fig/fig_teaser.tex
\begin{figure}[t]
  \centering
   \includegraphics[width=\linewidth]{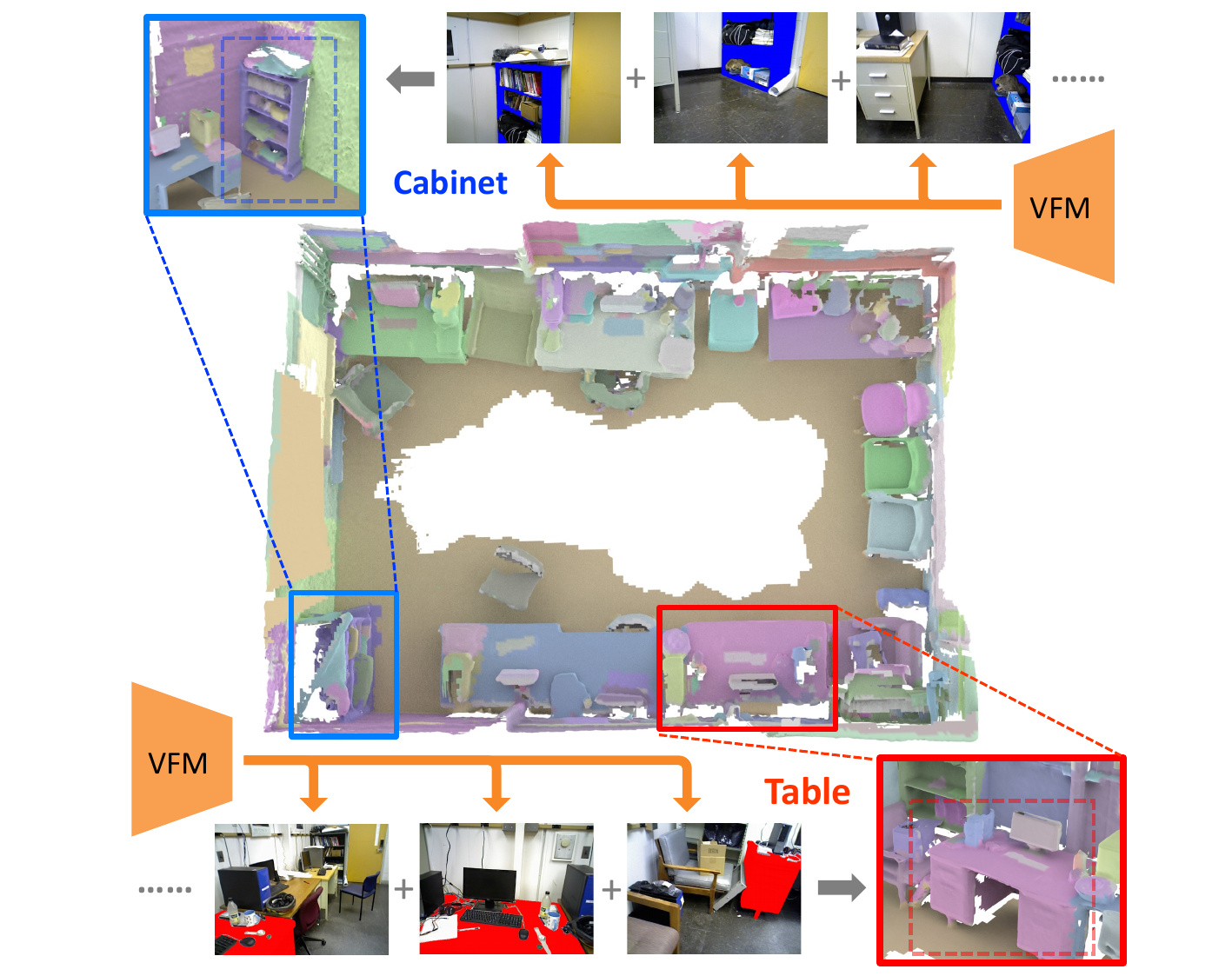}

   \caption{We propose an online zero-shot 3D segmentation method that establishes precise spatial associations between VFM-generated 2D masks from sequentially captured frames. We demonstrate an efficient merging process for masks detected from various viewpoints, enabling robust and consistent 3D instance segmentation in real time.}
   \label{fig:teaser}
   \vspace{-10pt}
\end{figure}

%% file: sec/2_related_works.tex
\section{Related Works}
\label{sec:related_works}

\paragraph{VFM for Offline 3D segmentation.}
Benefiting from the availability of vast amounts of 2D annotated data, many vision foundation models (VFMs)~\cite{kirillov2023segment, zhang2023simple, li2023semantic, zhao2023fast, lilanguage, zou2024segment, liu2024grounding} have developed rapidly in recent years, demonstrating strong capabilities and generalization across 2D segmentation tasks. However, high-quality 3D annotated data remains much more limited, significantly hindering the development of VFMs in 3D. As an alternative, researchers have turned to leveraging the power of 2D VFMs to assist with 3D segmentation tasks, exploring ways to bridge the gap between 2D and 3D visual understanding. 

With the assistance of VFMs, many methods have demonstrated surprisingly strong performance in 3D semantic and instance segmentation ~\cite{peng2023openscene, kerr2023lerf, huang2025openins3d, he2024unim, takmaz2023openmask3d, lu2023ovir, yan2024maskclustering}. They aim to transfer the knowledge learned from large-scale 2D datasets to 3D tasks by either aligning 3D points to 2D or back-projecting 2D information into 3D. 
In the first category, instances are detected directly in 3D space and projected into 2D pixel space~\cite{huang2025openins3d, takmaz2023openmask3d}. These projections are aligned with image pixels or regions to extract corresponding semantic features using VFMs. The aligned 2D pixel-level or region-level features are then aggregated in 3D space.
Conversely, methods in the second category focus on distilling 2D priors into 3D by back-projecting 2D information and evaluating spatial overlap relationships~\cite{yin2024sai3d, peng2023openscene, lu2023ovir, yan2024maskclustering, nguyen2024open3dis}. For example, Open3DIS~\cite{nguyen2024open3dis} using 2D generated masks to guide superpoint merging, while MaskClustering~\cite{yan2024maskclustering} leverages 2D segmentation from various viewpoints to detect spatially consistent 3D instances. These methods utilize the rich semantic information embedded in 2D images, transferring it to 3D by considering spatial overlaps, leading to more accurate and robust 3D instance segmentation.



\paragraph{Online 3D segmentation.}
With the rise of embodied AI and the growing demand for diverse robotic applications~\cite{chang2023context, li2024llm, koch2024open3dsg, zhang20233d}, online segmentation tasks have garnered increasing attention. Traditional online 3D segmentation methods typically rely on features extracted from sequentially acquired RGB-D frames using a pre-trained backbone~\cite{qi2017pointnet, qi2017pointnet++, ngo2023isbnet, choy20194d}, combined with feature aggregation techniques to achieve locally or globally consistent representations for final semantic predictions~\cite{zhang2020fusion, huang2021supervoxel, liu2022ins, weder2023alster, narita2019panopticfusion, mccormac2018fusion++, runz2018maskfusion, tateno2015real, grinvald2019volumetric}. While many of these methods have achieved impressive performance in closed-set settings through supervised training, they struggle to be extended to open-vocabulary settings, due to the limitation of 3D data.

To address this challenge by leveraging the broad knowledge of VFMs, a key obstacle lies in integrating the 2D predictions generated by VFMs from sequentially captured frames while maintaining real-time processing constraints. Some approaches attempt to distill semantic knowledge from sequential 2D inputs into a semantic 3D field in a frame-to-model manner~\cite{yamazaki2024open, tie20242, Schmid2024Khronos}. Additionally, another group of methods focus on instance-level information association. For instance, SAM3D~\cite{yang2023sam3d} processes sequential inputs in a bottom-up manner, while EmbodiedSAM~\cite{xu2024embodiedsam} trains a transformer-based model to support per-frame information merging in real time. Unlike these methods, our method performs online information merging primarily based on precise spatial associations between masks generated by VFMs, with feature similarities as auxiliary criteria.

%% file: sec/3_method.tex
\section{Method}
\label{sec:method}

\input{fig/fig_pipeline}

Given a stream of posed RGB-D frames $\{x_t=(C_t, D_t, T_t) | t=0, 1, ..., T\}$, where $C_t \in \mathbb{R}^{H \times W \times 3}$, $D_t \in \mathbb{R}^{H \times W}$ and $T_t \in \mathbb{R}^{4 \times 4}$ denote color image, depth image and camera pose respectively. Our goal is to segment all instances within the reconstructed 3D scene in an online manner. The output of our method includes the point cloud of reconstructed scene $S$, a set of 3D instance masks over $S$, and their corresponding open-vocabulary semantic features. 


\subsection{Overview}
\label{subsec:overview}
The overall pipeline of our method is illustrated in \cref{fig:pipeline}, which outlines the flow and key modules of our zero-shot online segmentation process. We employ a hashed voxel volume, denoted as $\mathit{Vol}$, for scene representation and maintain a mask bank $G$ to store extracted information of detected masks with spatial association. Each input frame is processed sequentially: first, it is integrated into $\mathit{Vol}$, and then its color image $C_t$ is fed to a pre-trained Visual Foundation Model (VFM) to generate 2D masks. Each detected 2D mask is subsequently lifted to a 3D mask through back-projection, and the corresponding hit voxels are extracted and inserted into the hash table (\cref{subsec:mask_scene_rep}) to label the overlapping associations. In parallel, relevant information of each mask is extracted and stored(\cref{subsec:mask_info_extract}).

As more 2D masks are detected from newly scanned frames, merging periodically the 2D masks belonging to the same 3D instance is necessary. The mask merging process is guided by their overlapping associations and feature similarity (\cref{subsec:online_mask_merging}). At the end of the input sequence, we can extract the reconstructed scene from the global volume, and each 3D instance's corresponding point cloud can also be accessed from the continuously updated hash table.

\subsection{Mask Bank with Spatial Associations}
\label{subsec:mask_scene_rep}
For an incoming frame $x_t=(C_t, D_t, T_t)$, we first adopt CropFormer~\cite{qi2023high} to generate entity-level 2D masks based on $C_t$. Each detected mask is then lifted into 3D through back-projection in $\mathit{Vol}$ assisted with depth image $D_t$ and frame pose $P_t$. For all $n_t$ masks detected up to timestamp $t$, we maintain a mask bank $G_t$ to efficiently store their key information, which is updated accordingly as masks are added or merged (\cref{subsec:online_mask_merging}).

\paragraph{Mask-Scene Association}
The primary task in dynamically maintaining the mask bank is to determine the spatial associations of masks across different frames within the 3D scene, enabling efficient overlap query between different masks. Similar to VoxelHashing~\cite{niessner2013real}, we represent the reconstructed scene as a hashed voxel volume $\mathit{Vol}$. Given a 3D coordinate of a certain voxel, the corresponding TSDF value can be directly queried in a hash table in $O(1)$. In addition to the TSDF value, each hash entry for a voxel $v_k$ maintains a list of masks' IDs that include $v_k$. Therefore, given a newly detected mask with $m$ voxels, all the masks associated spatially with it can be found in $O(m)$ and its ID is appended to the corresponding voxels then.


However, mask merge would lead to the frequent update of the mask ID list in each hash entry. To avoid the time overhead, we propose an append-only hash table update strategy. Specifically, instead of updating the mask ID in the hash table, the mask merge is updated in a mapping table. This table records the mapping between the original ID of each mask (which is stored in the hash table) and its updated ID. While two masks are merged in the following step, we just project their current IDs together to the same new one in this mapping table. Since the number of masks is significantly smaller than the number of their corresponding voxels, the time cost of the update caused by the merging can be ignored in this way.

\paragraph{Mask Representation in Database}
For all the detected masks, we record the following information in the database $M_t=\{ V_t, H_t, F^G_t, F^S_t, W_t, I_t \}$, where $V_t$ records each mask's corresponding voxels, and $H_t$ is the hash table at this timestamp. The semantic and geometric feature matrices, $F^S_t \in \mathbb{R}^{n_t \times d_s}$ and $F^M_t \in \mathbb{R}^{n_t \times d_g}$, store the semantic and geometric features of all masks, with $d_s$ and $d_g$ indicating the dimensionalities of the semantic and geometric features respectively. These features are critical to measure the similarity between different masks while merging.

Since the masks rarely merged with others are invalid with a higher possibility, we propose a mask weight value to indicate this characteristic. Each detected mask is initially assigned a weight of 1. When masks are merged, their weights are summed to determine the weight of the new mask. The weight of each mask is stored in the diagonal matrix $W_t=\operatorname{diag}(w_1, w_2, \dots, w_{n_t})$, where $w_i$ represents the weight of the $i$-th mask.

Additionally, $I_t$ records the \textbf{overlap ratio} of each pair of masks (introduced in \cref{subsec:mask_info_extract}), which indicates the spatial associations between masks and plays a significant role in the mask merging stage (\cref{subsec:online_mask_merging}). The value at position $[a,b]$ in $I_t$ represents the overlap ratio of the $a$-th mask to the $b$-th mask.

\subsection{Mask Merge Criteria}
\label{subsec:mask_info_extract}
To fuse all masks detected across different frames into 3D instances, the core of our online segmentation method involves dynamically recording and adjusting the associations between all masks, including their spatial overlap, semantic feature similarity, and geometric feature similarity, which serve as different criteria for mask merging, as detailed in \cref{subsec:online_mask_merging}. We first introduce the overlap ratio, which describes the degree of overlap between a pair of masks. Then, we introduce the extraction of both semantic features and geometric features for a mask.

\paragraph{Overlap ratio}
The key to evaluating the spatial associations between two masks lies in their degree of overlap. To quantify this, we propose to leverage the \textbf{overlap ratio}, as defined below, which can be computed online based on our voxel hashing-based scene representation.

Suppose that $X(m_i)$ represents the frames corresponding to the 2D masks that constitute the 3D mask $m_i$. Given any two masks in the mask bank, denoted as $m_a$ and $m_b$, with their corresponding voxel sets $V_a$ and $V_b$, and frame sets $X(m_a)$ and $X(m_b)$, our primary concern is the proportion of $m_a$ that includes $m_b$ (and vice versa). To compute it, first we need to identify the part of $m_b$ that is visible to $m_a$. This can be done by projecting all voxels in $V_b$ into the image planes of $X(m_a)$. The visible voxels are denoted as $\mathit{Vis}(V_b, X(m_a))=\{v_i \in V_b | v_i \rightarrow X(m_a) \}$. We can then compute their overlap by querying the hash table $H_t$ with $V_a$, yielding the intersection $V_a \cap V_b$. With this, the \textbf{overlap ratio} of $m_a$ to $m_b$ is defined as follows:
\vspace{-5pt}
\begin{equation}
    \mathit{or}_{(a,b)}=\frac{V_a \cap V_b}{\mathit{Vis}(V_b, X(m_a))}
    \label{eq:IR}
    \vspace{-5pt}
\end{equation}
This value quantifies the extent to which $m_a$ and $m_b$ occupy the same spatial position from the perspective of $m_a$.

\vspace{-10pt}
\paragraph{Semantic and Geometric Feature Extraction}
For a 2D mask generated by the VFM in frame $x_t$, its bounding box is cropped from $C_t$ at multiple scales and fed into CLIP~\cite{radford2021learning} to produce the open-vocabulary semantic feature.

For geometric feature extraction, the Marching Cubes algorithm~\cite{lorensen1998marching} is first applied to $\mathit{Vol}_{t}$ to obtain the so-far reconstructed scene point cloud $S_t$. This point cloud is then processed by FCGF~\cite{choy2019fully} to generate per-point features. Finally, we identify the points that lie within the voxel set of the mask and aggregate their point-wise geometric features using average pooling to obtain the final geometric feature for the mask.

\subsection{Online Mask Merging}
\label{subsec:online_mask_merging}
To obtain real-time 3D segmentation results and avoid the continuous increase in the number of detected masks, the online mask merging operation is applied to identify masks belonging to the same 3D instance and merge them into new masks. To fully leverage all available information up to the current timestamp $t$, our online merging strategy determines which masks should be merged based on the following criteria: (a) overlap ratio, (b) semantic similarity, (c) geometric similarity, and (d) consensus from third-view perspectives. In the following context, we first introduce our zero-shot online merging strategy, followed by the associated updating operations for the mask bank $G_t$.

\paragraph{Mask Merging Strategy}
With the correct spatial association between masks, determining whether two masks should be merged can be filtered by the similarity.
In general, we consider two masks to belong to the same 3D instance if either they are sufficiently similar overall, or there are enough third-view masks supporting their merging.

For the first criterion, we compute the overall similarity for all $n_t$ masks in $G_t$, incorporating their overlap ratio, semantic similarity, and geometric similarity. The Overall Similarity Matrix $\mathit{Sim}_t \in \mathbb{R}^{n_t \times n_t}$ is computed using the following formula:
\vspace{-5pt}
\begin{equation}
    \mathit{Sim}_t=\frac{1}{2}(\mathit{I}_t + \mathit{I}_t^\top) + F^S_t {F^S_t}^{\top} + F^G_t {F^G_t}^{\top}
    \label{eq:sim}
    \vspace{-5pt}
\end{equation}
where the first term represents the mutual overlap ratio between masks, while the second and third terms denote their semantic and geometric similarities respectively.

Additionally, we import the concept of "view consensus", adapted from MaskClustering~\cite{yan2024maskclustering} with some modifications to better fit the online task. For any two masks $m_a$ and $m_b$, if there exist another mask $m_c$ that satisfies the following conditions:
\vspace{-5pt}
\begin{align}
    ( or_{(c,a)} > \tau_1 ) \cap ( or_{(c,b)} > \tau_1 ) \label{eq:including} \\
    ( or_{(a,c)} > \tau_2 ) \cap ( or_{(b,c)} > \tau_2 ) \label{eq:included}
    \vspace{-5pt}
\end{align}
where $\tau_1=0.8$ is the threshold for inclusion, and $\tau_2=0.1$ is the threshold for being included. Then, $m_c$ is considered as a \emph{supporter} of $(m_a, m_b)$.

The condition in \cref{eq:including} indicates that from the perspective of $m_c$, both $m_a$ and $m_b$ are part of the same object, while the condition in \cref{eq:included} ensures that from both $m_a$ and $m_b$'s perspectives, $m_c$ is visible.

Notably, the supporter number matrix $A_t$, where the element at position $[a, b]$ denotes the supporter number of $(m_a,m_b)$, can be computed efficiently from $I_t$ and $W_t$ as follows:
\vspace{-5pt}
\begin{align}
    B &= ({I_t}^{\top} > \tau_1) \wedge (\mathit{I}_t > \tau_2) \vspace{-5pt}\\
    B' &= \text{ZeroDiag}(B) \vspace{-5pt}\\
    A_t &= B' W {B'}^{\top}
    \label{eq:supporter_num_mat}
    \vspace{-5pt}
\end{align}
where $B \in \mathbb{R}^{n_t \times n_t}$ is a binary matrix, and $B'$ is obtained by setting its diagonal elements to zero. The operator $\wedge$ denotes element-wise "and" operation between two matrices.

Combing the above two criteria, whether a pair of masks $(m_a, m_b)$ needs to be merged is evaluated by the following condition:
\vspace{-5pt}
\begin{equation}
    ( \mathit{Sim}_t[a,b] > \tau_{\text{sim}} ) \cup ( A_t[a,b] > \tau_{\text{supporter}} )
    \label{eq:merge}
    \vspace{-5pt}
\end{equation}
where $\tau_{\text{sim}}$ and $\tau_{\text{supporter}}$ are thresholds for overall similarity and supporter number respectively. 

\vspace{-10pt}

\paragraph{Updating the Mask Bank}
\label{subsubsec:merge_update}
\input{fig/fig_mapping_table}

After identifying the mask pairs that need to be merged, we group all mask pairs into clusters, where overlapping mask pairs are placed in the same cluster. Then we merge the masks within each cluster into a new mask, and the corresponding data structure in the mask bank $G_t$ is updated synchronously. At this stage, all masks in $G_t$ can now be categorized into two groups: retained masks (those that do not need merging) and combined masks (those that will be merged with others to form a new mask). The merging and updating process involves the following steps:

\begin{enumerate}
\item \textbf{Updating $\mathbf{V_t}$ and $\mathbf{W_t}$:} The corresponding voxel set of a new mask is formed by taking the union of the voxel sets of its constituent masks, and the weight of this new mask is the sum of their individual weights.

\item \textbf{Updating $\mathbf{F^G_t}$ and $\mathbf{F^S_t}$:} For each newly created mask, its semantic feature is aggregated from its constituent mask by average pooling, and its geometric feature is re-extracted using its updated voxel set from the latest reconstructed point cloud.

\item \textbf{Re-assigning global mask ID:} Each mask after merging, including both retained masks and new masks, is assigned a new global mask ID. We maintain a mapping table that tracks the original ID of each mask and its corresponding current global mask ID, and only this table is updated accordingly. This approach eliminates the need for frequent updates of the hash table $H_t$, significantly reducing time consumption. An illustration of this process is given in \cref{fig:mapping_table}.

\item \textbf{Updating $\mathbf{I_t}$}: Since all masks are reorganized, the overlap ratio matrix should be synchronized. Specifically:
(1) the rows and columns corresponding to remained masks stay unchanged, while those for the combined masks are removed. 
(2) rows and columns for new masks are then appended, and each element is recalculated by querying $H_t$ with its updated voxel set.
\end{enumerate}

\subsection{Implementation Details}
\label{subsec:imple_detail}
To ensure the efficiency of our method, we select keyframes at fixed intervals of 10 frames (or 20 frames for datasets with slow camera movement, such as SceneNN~\cite{hua2016scenenn}), with segmentation applied only to these keyframes. The mask merging operation is performed every 5 keyframes. At the end of the input sequence, we apply the same post-processing approach as in OVIR-3D~\cite{lu2023ovir}, to refine detected 3D instances.  For the final 3D instance prediction, only merged masks with weights greater than a threshold of $\tau_{\text{weight}}=5$ are considered valid instances and reported. The hyperparameters are set as $\tau_{\text{sim}}=2.3, \tau_{\text{supporter}}=5$. We test our method on an NVIDIA RTX 4090 GPU.

%% file: fig/fig_pipeline.tex
\begin{figure*}[t]
  \centering
   \includegraphics[width=\linewidth, trim=0 0 0 0, clip]{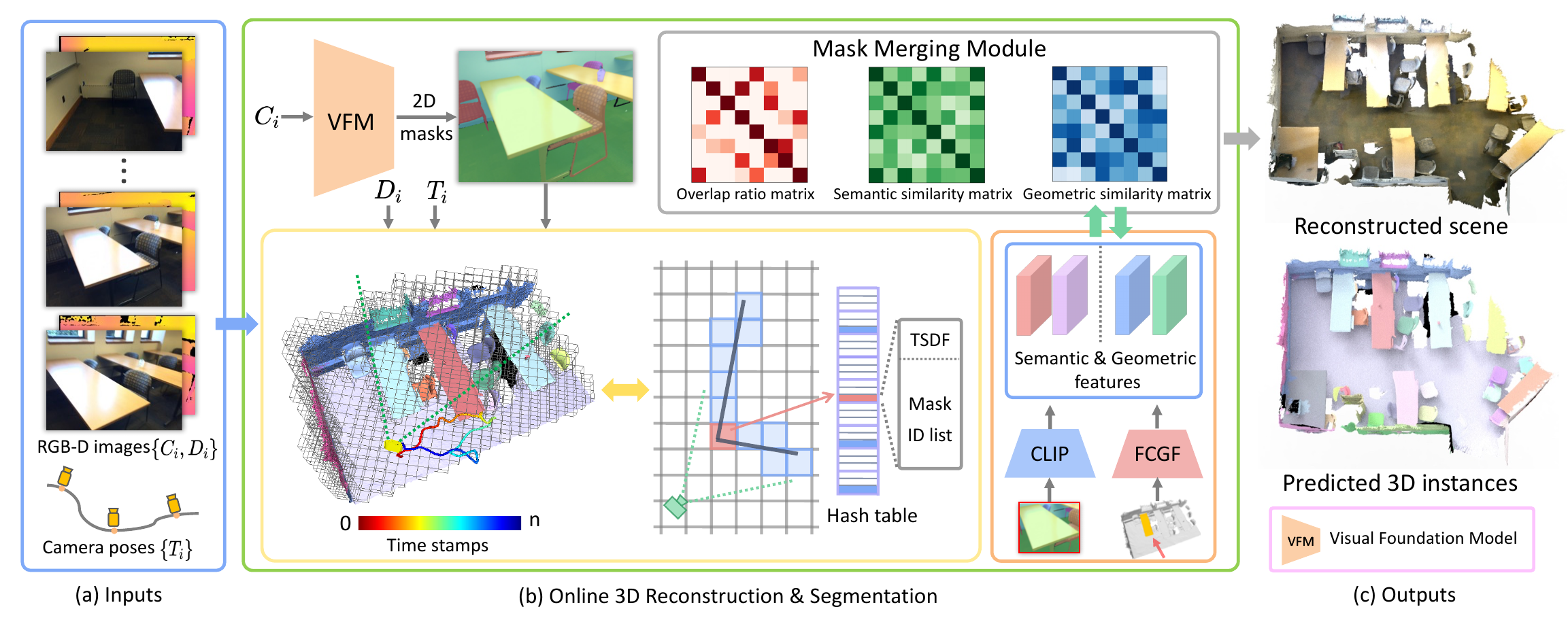}

   \caption{ \textbf{Overall pipeline.} (a) A posed RGB-D stream is input to our method sequentially. (b) A series of 2D masks are generated by VFM from the input color image and back-projected into 3D space, establishing associations with the VoxelHashing scene representation. Meanwhile, semantic and geometric features of the masks are extracted from pre-trained feature extractors and, together with mask overlap associations, serve as the core criteria for the Mask Merging process. (c) The final prediction of 3D instances is then output. }
   \label{fig:pipeline}

   \vspace{-10pt}
\end{figure*}

%% file: fig/fig_mapping_table.tex
\begin{figure}[t]
  \centering
   \includegraphics[width=\linewidth, trim=0 0 2.0 0, clip]{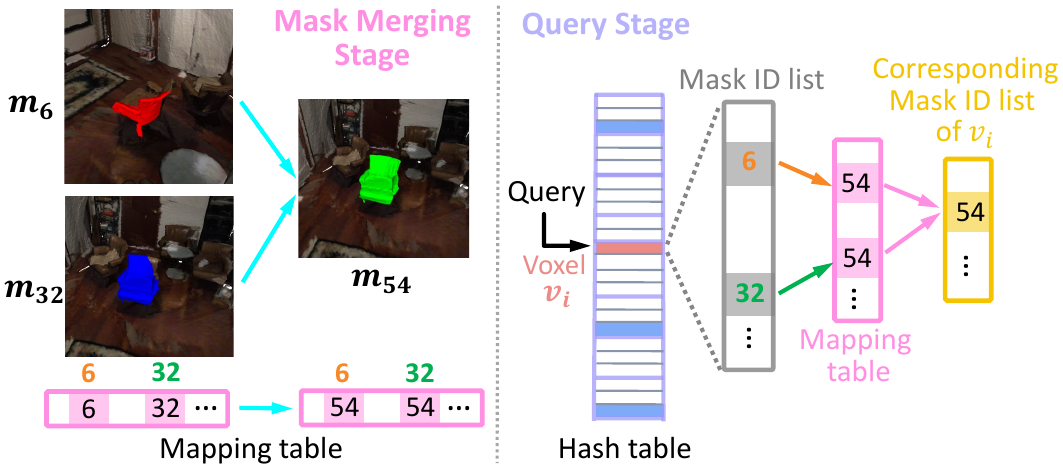}

   \caption{\textbf{The dynamically synchronized mapping table.} The mapping table is updated during the Mask Merging stage and facilitates efficient query in the Query stage, allowing the hash table to remain append-only. }
   \label{fig:mapping_table}
   \vspace{-10pt}
\end{figure}

%% file: sec/4_experiments.tex
\section{Experiments}
\label{sec:experiments}

\input{table/table_whole_seq}

In this section, we present extensive experiments to evaluate our method against state-of-the-art methods on publicly available datasets of 3D instance segmentation. 
We first introduce the experiment setup (\cref{subsec:experimental_setup}) and then report the quantitative experiments \cref{subsec:quantitative_results}  and qualitative results \cref{subsec:quantitative_results}. Finally, we conduct an ablation study to prove the effectiveness of our key designs \cref{subsec:ablation_study}.

\subsection{Experimental Setup}
\label{subsec:experimental_setup}
\paragraph{Datasets.} 
We conduct experiments on 3D instance segmentation benchmarks that contain real-world RGB-D datasets, including ScanNet200~\cite{dai2017scannet, rozenberszki2022language} and SceneNN~\cite{hua2016scenenn}. (1) \textbf{ScanNet200} is an indoor dataset comprising 1513 room-level sequences, each annotated with instance-level segmentation and labels across 200 categories. Consistent with the comparison methods, we evaluate our approach on the validation set, which includes 312 sequences. (2) \textbf{SceneNN} contains over 100 indoor scenes with instance-level segmentation annotations. Following EmbodiedSAM~\cite{xu2024embodiedsam}, we adopt the same 12 high-quality scenes for evaluation. We report the scene ID in the Supplementary.

\paragraph{Baselines.}

We compare our method with both offline methods and online methods. For offline methods, we choose recent advanced works including OVIR-3D~\cite{lu2023ovir} and MaskClustering~\cite{yan2024maskclustering}, which are both fully zero-shot offline segmentation methods. 


For online methods, we compared our method with recent works SAM3D~\cite{yang2023sam3d} and EmbodiedSAM~\cite{xu2024embodiedsam}. SAM3D is a zero-shot segmentation method, which sequentially processes the input sequence and merges the segmentation in a bottom-up manner.  EmbodiedSAM~\cite{xu2024embodiedsam} trains a transformer to learn the merging process between incoming masks generated by the VFM. Unlike our zero-shot fashion, the merging operation in EmbodiedSAM needs to be trained on the ScanNet200 dataset.


\paragraph{Metrics.}
Following previous works~\cite{yan2024maskclustering}, we employ the standard Average Precision (AP) metric under IoU thresholds of 25\% and 50\%, as well as the mean AP across IoU thresholds from 50\% to 95\%, denoted as $AP_{25}$, $AP_{50}$ and $AP$ respectively. This metric evaluates the overall accuracy of all predicted instances against all ground truth instances. For simplicity, percentage signs are omitted from all reported values in the following tables.

\subsection{Quantitative Results}
\label{subsec:quantitative_results}
\input{table/tabel_intermediate_results3}

\paragraph{Full-sequence Segmentation Results.} 
We evaluate the full-sequence segmentation results on both offline and online segmentation methods on ScanNet200 and SceneNN, with results presented in \cref{tab:whole_seq}. Compared to the zero-shot offline segmentation method MaskClustering, our method achieves comparable performance on ScanNet200. Notably, our method can even outperform MaskClustering on more challenging SceneNN sequences. Compared to the online method SAM3D which leverages the same setup as ours, our method yields an approximate +9\% improvement in $AP$.
EmbodiedSAM achieves the best performance on ScanNet200 since it was trained on this dataset. However, a significant performance drop is observed in the evaluation on SceneNN, which demonstrates that the supervised learning approach for mask merging lacks generalizability. 
Besides, our method achieves the highest running efficiency (improved from 10 FPS to 15 FPS ) during segmentation. 



\paragraph{Intermediate-sequence Segmentation Results.} 
To assess online performance during scanning, we evaluate segmentation outcomes on partially scanned sequences, specifically at $25\%$, $50\%$, and $75\%$ completion, with no post-processing applied. These intermediate sequences introduce reconstruction noise and substantial occlusions, as shown in \cref{tab:inter_result}. Our method demonstrates significantly improved results over other baselines, including EmbodiedSAM, which was trained on ScanNet200, achieving gains of approximately +4\%, +8\%, and +10\% in $AP$, $AP_{50}$, and $AP_{25}$, respectively. These findings highlight the robustness of our approach.



\subsection{Qualitative Results}
\label{subsec:qualitative_results}
\input{fig/fig_visual_realtime_seg}
\input{fig/fig_visual_query}

\input{fig/fig_visual_comp}

In \cref{fig:visual_realtime_seg}, we present the intermediate segmentation results of the online methods, directly taken during the sequence scanning process and displayed on the reconstructed mesh or point cloud output by each method, without any post-processing. Compared to EmbodiedSAM, our real-time segmentation results exhibit significantly less noise, which can be attributed to our mask merging strategy that effectively utilizes global mask information to guide the merging process. Additionally, some visual examples of real-time open-vocabulary querying are provided in \cref{fig:visual_query}. This process is implemented by encoding the query text into embeddings and computing similarity to the aggregated semantic features of the currently detected valid instances.

We also provide a visual comparison of our method with other baseline methods on the ScanNet200 and SceneNN in \cref{fig:visual_comp}. For offline methods, segmentation results are directly displayed on the input ground truth mesh, while for online methods, the results are mapped from the output mesh or point cloud to the ground truth mesh using point correspondences to ensure a fair comparison. EmbodiedSAM performs well on ScanNet200, where it was trained, but experiences a performance drop on SceneNN. As a zero-shot segmentation method, our method demonstrates greater stability across different datasets and achieves performance comparable to MaskClustering, the SOTA offline zero-shot method.

\subsection{Ablation Study}
\label{subsec:ablation_study}

\input{table/table_ablation_module}

Since the mask merging strategy is the key technique of our method, we conduct experiments to evaluate the effectiveness of various criteria as shown in \cref{tab:ablation_criteria}. 
When retaining only the overlap ratio and third-view supporting criteria, we observe an AP drop of approximately 8\%, while relying solely on feature similarities leads to a substantial 50\% decrease in AP.
These findings reveal that criteria related to spatial associations are most critical, underscoring the importance of precise spatial alignment in merging 2D segmentation results into 3D. Conversely, relying exclusively on semantic and geometric features from standard feature extractors results in poor performance. Such locally consistent features lack global distinctiveness, causing objects with similar local shapes to be mistakenly identified as identical, regardless of their spatial associations.


\vspace{-2pt}
\subsection{Limitations}
\label{subsec:limitations}
While our method demonstrates strong performance, there are several notable limitations. First, although our merging-based strategy effectively combines masks from various viewpoints and manages over-segmented masks, it is less robust in handling under-segmentation, which can reduce accuracy in scenes with significant occlusions. Furthermore, since our method fundamentally follows a space-time tradeoff strategy, it encounters challenges in scaling to very large environments, such as floor-level scenes.


%% file: table/table_whole_seq.tex
\begin{table*}[ht]

\centering
\small  
\setlength{\tabcolsep}{8.5pt}  
\renewcommand{\arraystretch}{1.1}  

\begin{tabular}{l|cc|ccc|ccc|c}
\specialrule{1.0pt}{0pt}{0pt}  
\multirow{2}*{\textbf{Method}} & \multirow{2}*{\textbf{Online}} & \multirow{2}*{\textbf{Zero-shot}} & \multicolumn{3}{c|}{\textbf{ScanNet200}} & \multicolumn{3}{c|}{\textbf{SceneNN}} & FPS \\
~ & ~ & ~ & $AP$ & $AP_{50}$ & $AP_{25}$ & $AP$ & $AP_{50}$ & $AP_{25}$ &  \\
\specialrule{1.0pt}{0pt}{0pt}
EmbodiedSAM~\cite{xu2024embodiedsam} & \ding{51} & \ding{55} & \textbf{28.8} & \textbf{42.7} & \textbf{54.2} & \textbf{20.1} & 32.5 & 46.3 & 10 \\
OVIR-3D~\cite{lu2023ovir} & \ding{55} & \ding{51} & 14.4 & 27.5 & 38.8 & 12.3 & 24.4 & 34.6 & - \\
MaskClustering~\cite{yan2024maskclustering} & \ding{55} & \ding{51} & 19.7 & 36.4 & 51.4 & 16.3 & 31.7 & 46.2 & - \\
SAM3D~\cite{yang2023sam3d}$^*$$\ddag$ & \ding{55} & \ding{51} & 17.8 & 30.6 & 48.5 & - & - & - & 8 \\
\hline
SAM3D~\cite{yang2023sam3d}$^*\dag$ & \ding{51} & \ding{51} & 9.6 & 24.8 & 49.6 & 9.1 & 21.3 & 43.4 & 8 \\
Ours & \ding{51} & \ding{51} & 18.6 & 36.1 & 53.5 & 18.1 & \textbf{35.3} & \textbf{59.5} & \textbf{15} \\
\specialrule{1.0pt}{0pt}{0pt}  
\end{tabular}

\caption{\textbf{Full-sequence instance segmentation results on ScanNet200 and SceneNN.} For the online methods, the instance segmentation results are mapped from their reconstructed point cloud or mesh to ground truth point cloud through point correspondences. $^*\dag$: Raw outputs generated by SAM3D, $^*\ddag$: Ensembled outputs~\cite{yang2023sam3d}, raw outputs merged with other over-segmentation results. }

\label{tab:whole_seq}
\end{table*}

%% file: table/tabel_intermediate_results3.tex
\begin{table*}[ht]

\centering
\small  
\setlength{\tabcolsep}{4pt}  
\renewcommand{\arraystretch}{1.25}  

\begin{tabular}{l|cc|ccc|ccc|ccc|ccc}
\specialrule{1.0pt}{0pt}{0pt}  

\multirow{2}*{\textbf{Method}} & \multirow{2}*{\textbf{Online}} & \multirow{2}*{\textbf{Zero-shot}} & \multicolumn{3}{c|}{\textbf{25\%}} & \multicolumn{3}{c|}{\textbf{50\%}} & \multicolumn{3}{c|}{\textbf{75\%}} & \multicolumn{3}{c}{\textbf{Final}}\\

~ & ~ & ~ & $AP$ & $AP_{50}$ & $AP_{25}$ & $AP$ & $AP_{50}$ & $AP_{25}$ & $AP$ & $AP_{50}$ & $AP_{25}$ & $AP$ & $AP_{50}$ & $AP_{25}$ \\
\specialrule{1.0pt}{0pt}{0pt}

SAM3D~\cite{yang2023sam3d}$^*\dag$ & \ding{51} & \ding{51} & 9.7 & 22.5 & 41.8 & 8.8 & 23.8 & 44.5 & 10.0 & 23.2 & 42.4 & 9.1 & 21.3 & 43.4 \\
EmbodiedSAM~\cite{xu2024embodiedsam} & \ding{51} & \ding{55} & 12.1 & 28.5 & 48.6 & 11.8 & 28.6 & 48.1 & 12.2 & 26.9 & 50.4 & \textbf{20.1} & 32.5 & 46.3 \\
Ours & \ding{51} & \ding{51} & \textbf{18.0} & \textbf{36.8} & \textbf{59.3} & \textbf{17.4} & \textbf{36.7} & \textbf{60.7} & \textbf{18.3} & \textbf{35.8} & \textbf{58.8} & 18.1 & \textbf{35.3} & \textbf{59.5} \\

\specialrule{1.0pt}{0pt}{0pt}  
\end{tabular}

\caption{ \textbf{Intermediate instance segmentation results of the online methods on SceneNN Dataset.} For example, \textbf{25\%} in table represents the intermediate segmentation results after 25\% of the input sequence has been processed. $^*\dag$: Raw outputs generated by SAM3D~\cite{yang2023sam3d} }
\label{tab:inter_result}

\vspace{-10pt}
\end{table*}

%% file: fig/fig_visual_realtime_seg.tex
\begin{figure}[t]
  \centering
   \includegraphics[width=\linewidth, trim=0 0 0 0, clip]{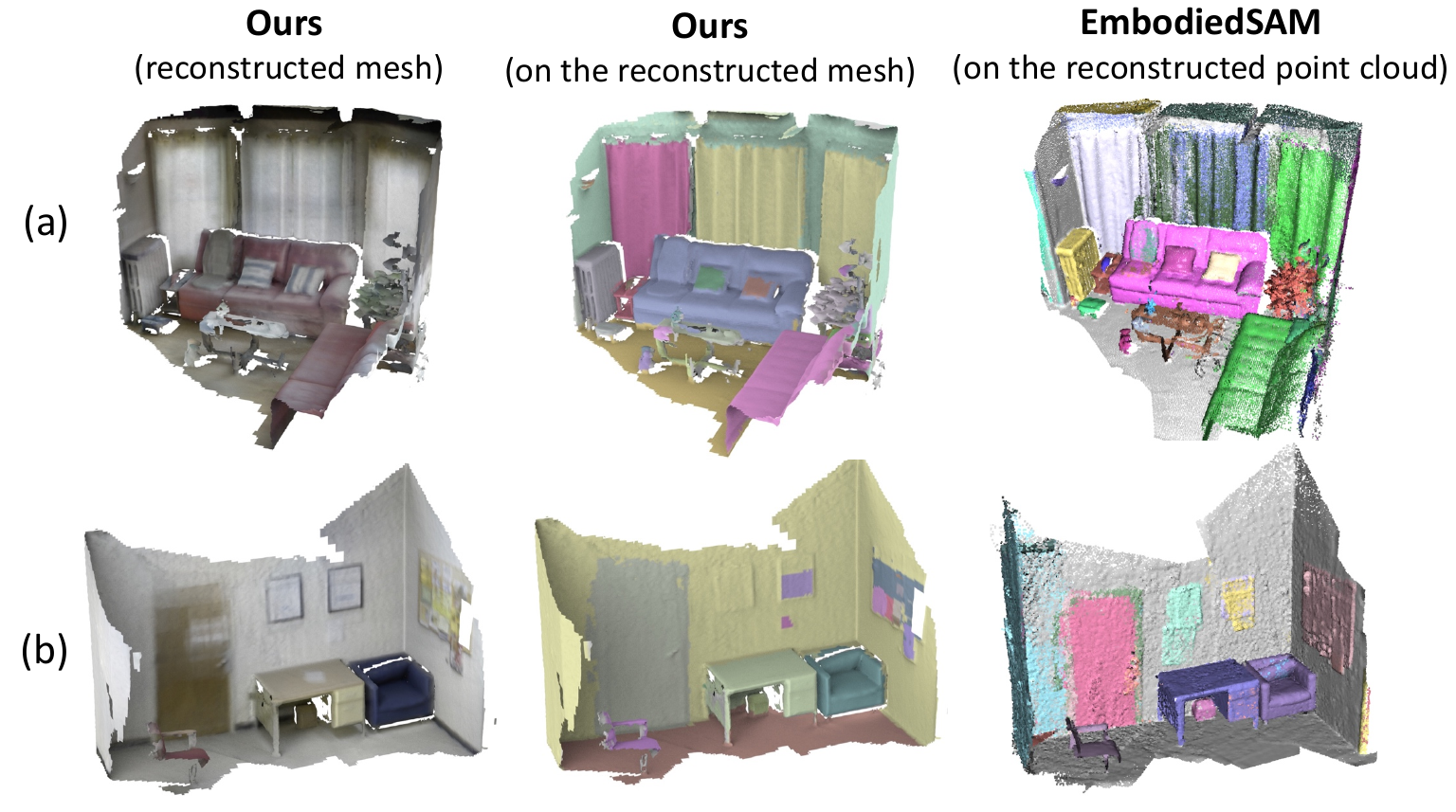}

   \caption{Intermediate instance segmentation results, displayed on each method’s reconstructed mesh or point cloud.}
   \label{fig:visual_realtime_seg}
   \vspace{-10pt}
\end{figure}

%% file: fig/fig_visual_query.tex
\begin{figure}[ht]
  \centering
   \includegraphics[width=\linewidth, trim=0 0 0 0, clip]{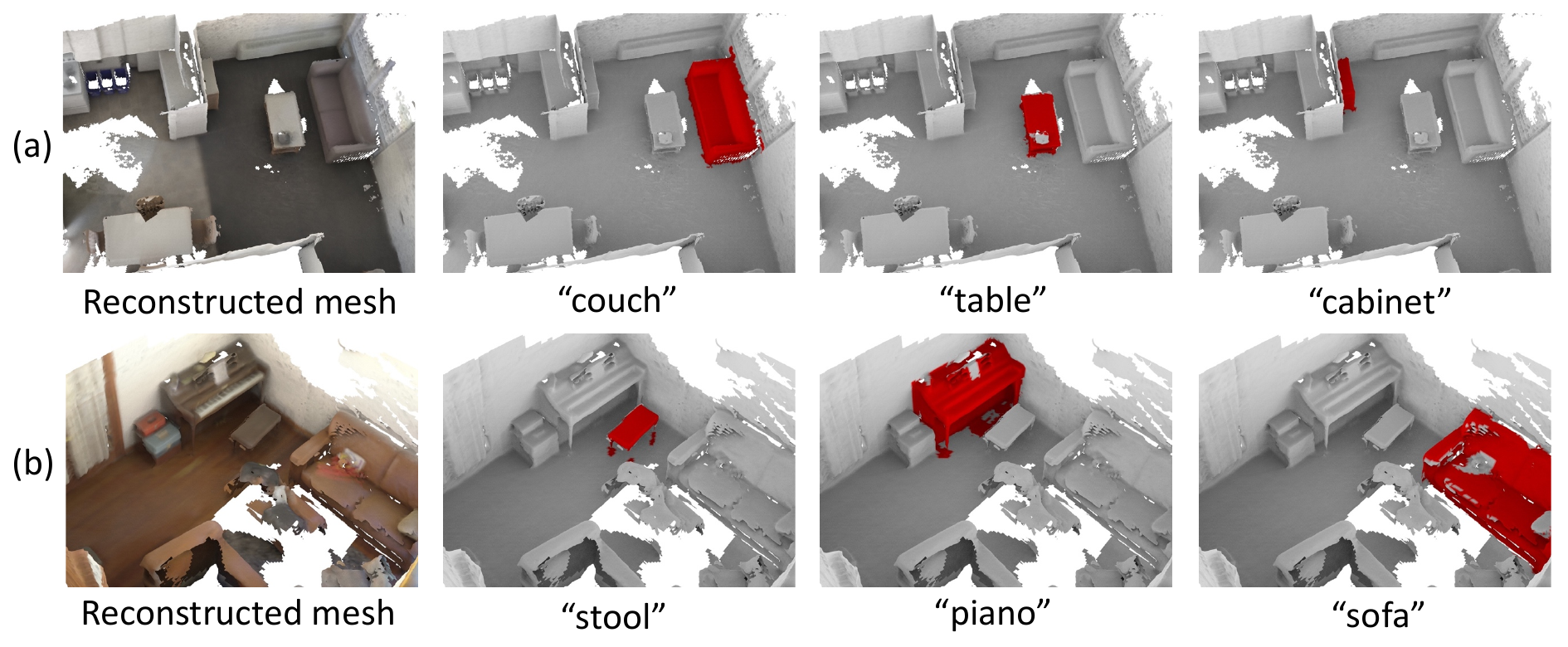}

   \caption{Open-vocabulary instance retrieval with varied query texts during the scanning process.}
   \label{fig:visual_query}
   
   \vspace{-10pt}
\end{figure}

%% file: fig/fig_visual_comp.tex
\begin{figure*}[h]
  \centering
   \includegraphics[width=0.995\linewidth, trim=2.0 0.5 0 7.0, clip]{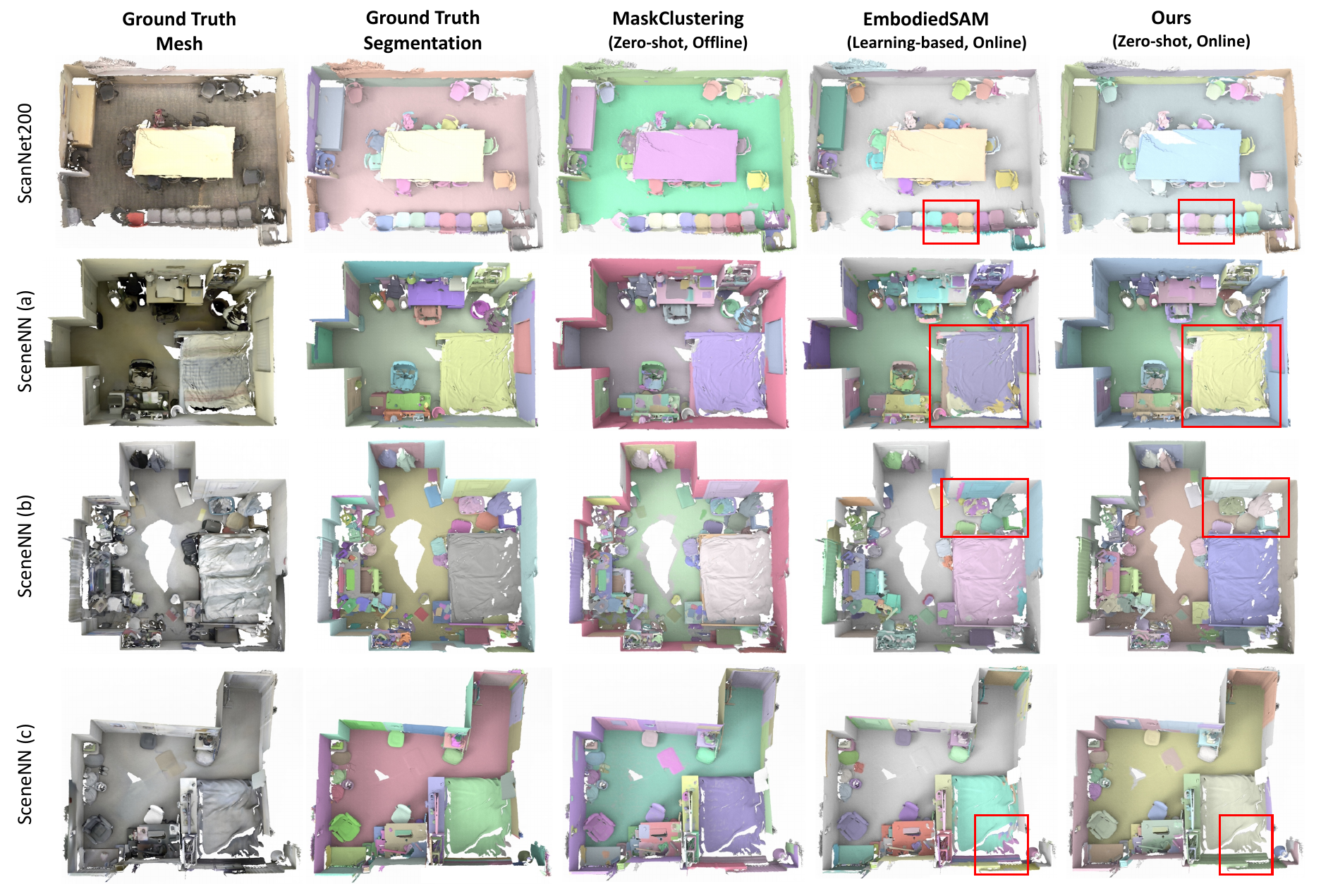}
   \vspace{-2pt}

   \caption{Comparison of segmentation results on full-sequences with other SOTA methods, including the zero-shot offline method MaskClustering~\cite{yan2024maskclustering} and the learning-based online method EmbodiedSAM~\cite{xu2024embodiedsam}. Background regions are shaded in gray. The examples demonstrate that our method achieves a more accurate merge of 2D masks, significantly reducing noise in the segmentation results.}
   \label{fig:visual_comp}

   \vspace{-10pt}
\end{figure*}

%% file: table/table_ablation_module.tex
\begin{table}[h]
\vspace{-8pt}

\centering
\small
\setlength{\tabcolsep}{5pt}  
\renewcommand{\arraystretch}{1.2}  

\begin{tabular}{l|ccc}
\specialrule{1.0pt}{0pt}{0pt}  
  & $AP$ & $AP_{50}$ & $AP_{25}$ \\
\specialrule{1.0pt}{0pt}{0pt}
\vspace{-2pt}


Only feature similarity & 9.7 & 19.7 & 36.9 \vspace{-1pt} \\
No overlap ratio & 13.7 & 26.1 & 43.1 \vspace{-1pt} \\
No third-view supporting & 16.9 & 30.1 & 46.3 \vspace{-1pt} \\
No feature similarity & 17.1 & 33.0 & 49.4  \vspace{-1pt} \\
Full merging strategy & \textbf{18.6} & \textbf{36.1} & \textbf{53.5} \\

\specialrule{1.0pt}{0pt}{0pt}  
\end{tabular}

\caption{Ablation study on different criteria in our online merging strategy on ScanNet200. Notably, the spatial associations play the most crucial role in achieving accurate mask merging results.}
\label{tab:ablation_criteria}

\vspace{-7pt}
\end{table}

%% file: sec/5_conclusion.tex
\section{Conclusion}
\label{sec:conclusion}

In this work, we present OnlineAnySeg, a straightforward yet effective approach for online organization and merging of instance masks provided by vision foundation models, using a hashing technique. We propose lifting predicted, inconsistent 2D masks into 3D based on their spatial associations, using a similarity-based filtering strategy to accurately generate 3D instance masks in a zero-shot manner. By leveraging voxel hashing for efficient 3D scene query, we reduce the time complexity of the costly spatial overlap query from $O(n^2)$ to $O(n)$ compared to the pairwise mask association strategy, making this the first method to effectively use explicit spatial associations to enhance segmentation performance under real-time constraints. This design allows mask merging to be free from the constraints of the limited training data distribution, making our approach more robust to incomplete and noisy data. Experimental results on datasets like SceneNN demonstrate that our approach offers a clear accuracy advantage over other online methods when applied to incrementally scanned data while achieving the highest efficiency. Moreover, our method attains results comparable to offline approaches. We hope our method inspires future work to explore lifting 2D predictions from VFMs to tackle more complex 3D tasks.

%% file: sec/6_acknowledgements.tex
\section{Acknowledgements}
\label{sec:acknowledgements}

This work was supported in part by the NSFC (62325211, 62132021, 62372457) and the Major Program of Xiangjiang Laboratory (23XJ01009).

%% file: sec/X_suppl_1_main.tex
\clearpage
\setcounter{page}{1}
\maketitlesupplementary


\section{Overview}
In the supplementary material, the sections are briefly introduced as follows:

\begin{itemize}
\item We provide more detailed analyses for certain modules of our method in \cref{sec:suppl_method_details}. 

\item Additional experimental results, including both quantitative and qualitative analyses, are presented in \cref{sec:suppl_add_exp}.

\item In \cref{sec:suppl_video} we present a video demo showcasing the process of our online segmentation as well as visual comparisons to other online segmentation methods.

\end{itemize}

\input{sec/X_suppl_2_method_detail}

\input{sec/X_suppl_3_add_exp}

\input{sec/X_suppl_4_video}

%% file: sec/X_suppl_2_method_detail.tex
\section{Method Details}
\label{sec:suppl_method_details}

\subsection{Discussion about Overlap Ratio}
As introduced in Sec. 3.3 of the main paper, we define the \textbf{Overlap Ratio}, which quantifies the overlap between a pair of masks in 3D space, with the overlap ratios for all mask pairs stored in the matrix $I_t$. Another straightforward way to measure the overlap between two masks $m_a$ and $m_b$ is to calculate the proportion of overlapping voxels relative to the voxel count of the mask, for example: 
\begin{equation}
    \mathit{or*}_{(a,b)}=\frac{|V_a \cap V_b|}{|V_b|}
    \label{eq:supp_or}
\end{equation}
without considering the visual part of the visible part of mask $m_b$ in the frame set of mask $m_a$. However, we find that this approach is not feasible for our method. A detailed analysis is provided here.

\input{fig/fig_suppl_overlap_ratio}

An example of two masks belonging to the same 3D instance is given in \cref{fig:suppl_overlap_ratio}. These masks represent observations of the same chair, captured from different viewpoints (as highlighted by the green bounding box). The voxel sizes of the lifted 3D masks are 3520 and 644 respectively, with an intersection size of 358 voxels. According to the definition in \cref{eq:supp_or}, the overlap ratio of $m_a$ to $m_b$ is calculated as $or^{*}_{(a,b)}=\frac{358}{644}=0.556$, which is considerably lower than what intuition might suggest. This is because, with the significant difference between the two viewpoints, the two masks cannot align perfectly in 3D space, due to depth noise and occlusion. On the other hand, if we set the threshold for the overlap ratio too low (e.g., 0.5), it may lead to other problems. For instance, if a chair is located very close to a table, and a mis-segmentation occurs in a 2D frame where a large portion of the chair is incorrectly segmented as part of the table, the merging strategy may erroneously combine them into a single instance, making our method highly sensitive to under-segmentation.

By incorporating the "visible part", the overlap ratio of $m_a$ to $m_b$ is $or_{(a,b)} = 1.0$ (following the definition of the overlap ratio in the main paper Sec. 3.3), which aligns with their identity in 3D space. A qualitative comparison of the two different methods for calculating the overlap ratio is provided in \cref{tab:suppl_overlap_ratio}.

\input{table/table_suppl_overlap_ratio}

\subsection{Extraction of Geometric Feature}

\input{fig/fig_suppl_geo_feature_pc}
In Sec. 3.3 of the main paper, we describe the extraction of geometric features for each detected mask using FCGF~\cite{choy2019fully}. A visualization of the extracted feature point clouds by FCGF is presented in \cref{fig:suppl_geo_feature_pc}, where points with similar colors indicate higher feature similarity. In our method, the per-point geometric features of the latest reconstructed point cloud $S_t$ are extracted, and the features for each mask are aggregated based on their corresponding scene points. as illustrated in ~\cref{fig:suppl_geo_feature} (b). Compared to the naive method of directly feeding the back-projected point cloud of each detected mask into the geometric feature extractor, as shown in ~\cref{fig:suppl_geo_feature} (a), our approach is both more accurate and time-efficient.

\input{fig/fig_suppl_geo_feature}

First, as an input sequence typically contains thousands of masks in total, extracting geometric features for each mask individually can result in significant time overhead. 
Additionally, since FCGF is a fully convolutional network capable of capturing broad spatial context, using a more complete input point cloud produces higher-quality output features. Visualizations of the output feature point clouds are presented in ~\cref{fig:suppl_geo_feature}, where more similar colors indicate higher feature similarity. For a pair of masks $(m_a, m_b)$ corresponding to the same 3D instance observed from different viewpoints (denoted as frame 1 and frame 2), the significant disparity between the viewpoints can lead to notable feature dissimilarities when extracted separately (~\cref{fig:suppl_geo_feature}, a). In contrast, cropping the complete feature point cloud of $S_t$ ensures globally consistent features for the masks (~\cref{fig:suppl_geo_feature}, b).

\vspace{-10pt}
\subsection{Comparison with Frame-by-frame Mask Merging Strategy}
\label{subsec:supp_comp_fbf}

We describe our online mask merging strategy in Sec. 3.4 of the main paper, which establishes mask associations by overall similarities and third-view support. In contrast, some previous methods adopt a "frame-by-frame" merging strategy to process sequential inputs. For example, OVIR-3D~\cite{lu2023ovir} focuses on finding instance correspondences between newly detected masks in incoming frames and all existing masks from previous frames. Even though various post-processing operations can be applied to remove redundant instances, the "frame-by-frame" merging strategy can lead to significant issues, particularly in the following scenarios. A typical example is shown in \cref{fig:suppl_merge}.

\input{fig/fig_suppl_merge}

First, for a large object observed partially from different viewpoints with minimal overlap, the "frame-by-frame" merging strategy struggles to correctly associate the newly observed part with previous observations. For instance, as illustrated in \cref{fig:suppl_merge}, a long dining table is segmented into two separate instances by the "frame-by-frame" merging strategy. Second, if a previously detected object is scanned again from a completely different viewpoint, the newly detected instance is often not successfully matched to the existing one, resulting in it being treated as a new instance. This issue is exemplified by the chairs in \cref{fig:suppl_merge}.

Our mask merging strategy, on the other hand, fully leverages all previous observations to ensure global consistency. Additionally, by incorporating third-view supporting to establish extra associations between masks with limited overlap, these issues can be significantly mitigated. A qualitative comparison of these two merging strategies is presented in \cref{tab:suppl_merge}.

\input{table/table_suppl_merge}

\subsection{Comparison without using the mapping table}
\label{subsec:supp_comp_mt}
We introduce the mapping table in Sec. 3.2 of the main paper, which maps the IDs of the original 2D masks to the IDs of the current 3D masks, enabling the tracking of each mask throughout every merging process. Instead of directly updating the mask ID lists in the hashed voxel volume, the re-assignment of mask IDs during the merging stage (Section 3.4 of the main paper) only triggers a synchronous update of the mapping table, keeping the hash table append-only. This design significantly accelerates the updating of the mask bank $G_t$, as frequent voxel modifications in the volume are highly time-consuming.

A speed comparison of the scanning process with and without the mapping table is shown in \cref{fig:suppl_mappingtable}, demonstrating the effectiveness of this design. Without the mapping table, the process is approximately 10 times slower, and the slowdown can even reach up to 20 times as the number of masks increases.

\input{fig/fig_suppl_mappingtable}

%% file: fig/fig_suppl_overlap_ratio.tex
\begin{figure}[ht]
  \centering
   \includegraphics[width=\linewidth, trim=0 0 0 0, clip]{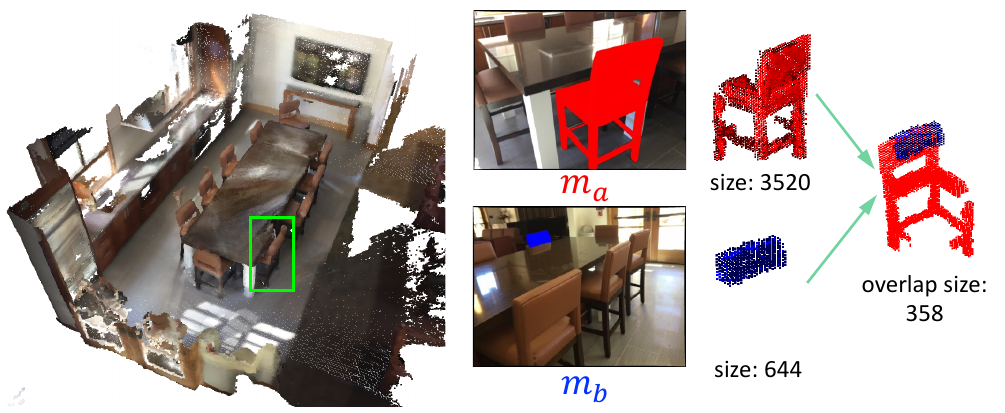}

   \vspace{10pt}
   \caption{An example of computing overlap ratio without considering "visible part".}
   \label{fig:suppl_overlap_ratio}
\end{figure}

%% file: table/table_suppl_overlap_ratio.tex
\begin{table}[h]

\centering
\setlength{\tabcolsep}{5pt}  
\renewcommand{\arraystretch}{1.2}  

\begin{tabular}{c|ccc}
\specialrule{1.0pt}{0pt}{0pt}  
  & $AP$ & $AP_{50}$ & $AP_{25}$ \\
\specialrule{1.0pt}{0pt}{0pt}

w/o "Visible Part"  & 13.6 & 26.9 & 40.3 \\
w   "Visible Part" & \textbf{18.6} & \textbf{36.1} & \textbf{53.5} \\

\specialrule{1.0pt}{0pt}{0pt}  
\end{tabular}

\vspace{10pt}
\caption{Comparison of different methods for calculating the \textbf{Overlap Ratio}.}
\label{tab:suppl_overlap_ratio}

\end{table}

%% file: fig/fig_suppl_geo_feature_pc.tex
\begin{figure*}[ht]
  \centering
   \includegraphics[width=\linewidth, trim=0 0 0 0, clip]{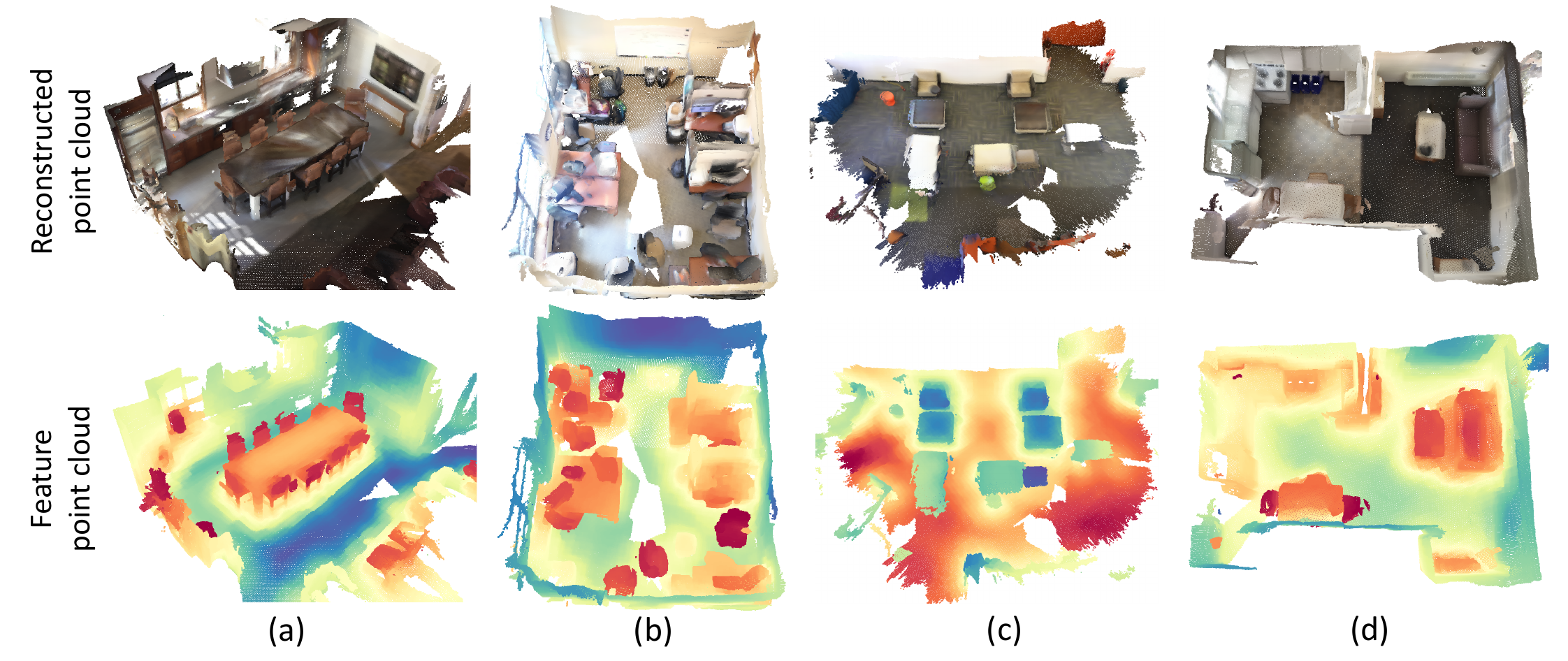}

   \caption{\textbf{Visualizations of feature point clouds output by FCGF~\cite{choy2019fully}.} The first row shows the reconstructed point clouds from our method. The second row displays their corresponding feature point clouds, colorized based on the extracted per-point features, where points with similar colors indicate high feature similarity within the same feature point cloud.}
   \label{fig:suppl_geo_feature_pc}
\end{figure*}

%% file: fig/fig_suppl_geo_feature.tex
\begin{figure}[ht]
  \centering
   \includegraphics[width=\linewidth, trim=0 0 0 0, clip]{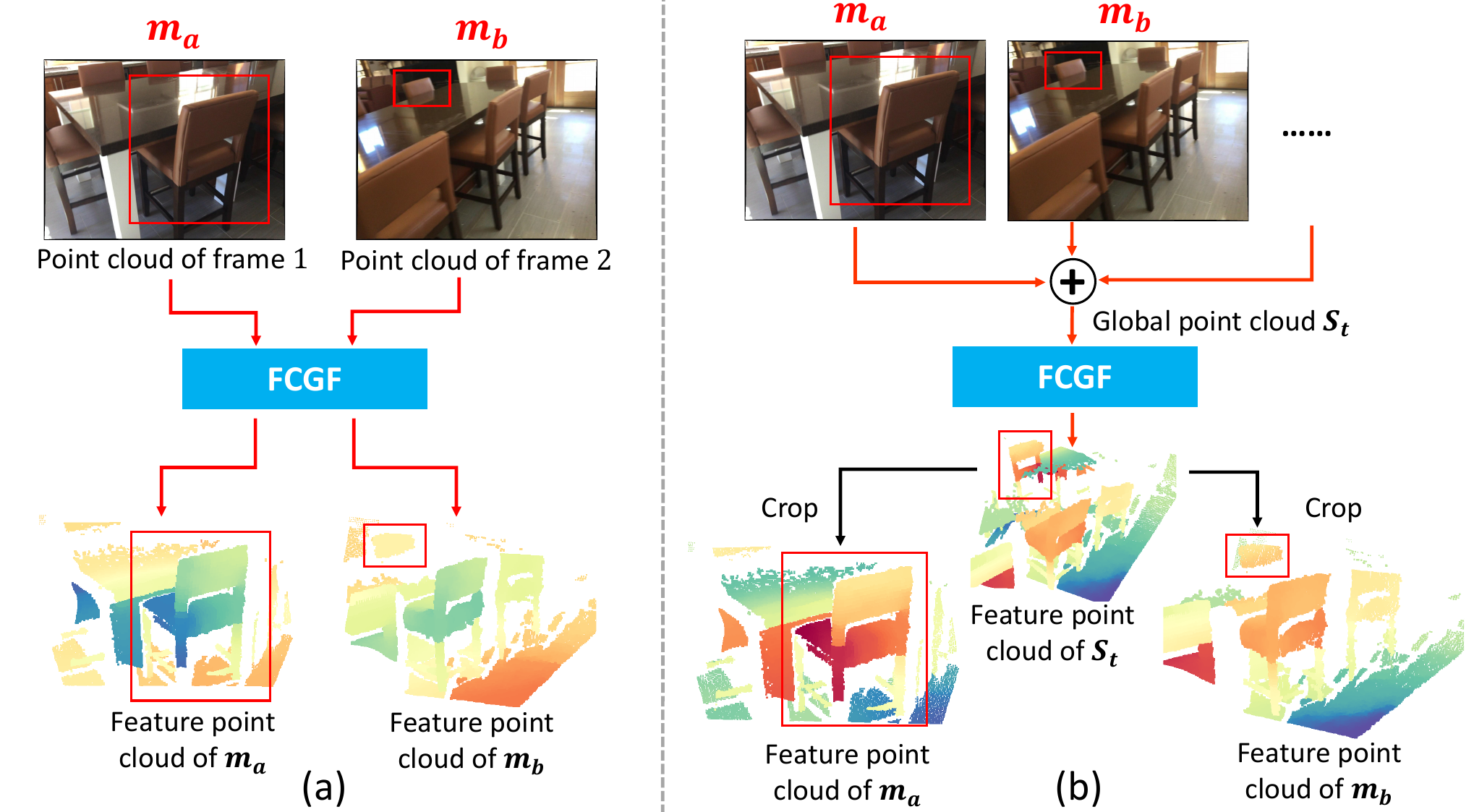}

   \vspace{10pt}
   \caption{\textbf{Comparison of different approaches for extracting geometric features of masks.} (a) Extracting point features separately for each point cloud. (b) Extracting point features from the global point cloud and cropping the resulting feature point cloud to obtain sub-feature point clouds.}
   \label{fig:suppl_geo_feature}
\end{figure}

%% file: fig/fig_suppl_merge.tex
\begin{figure}[h]
  \centering
   \includegraphics[width=\linewidth, trim=0 0 0 0, clip]{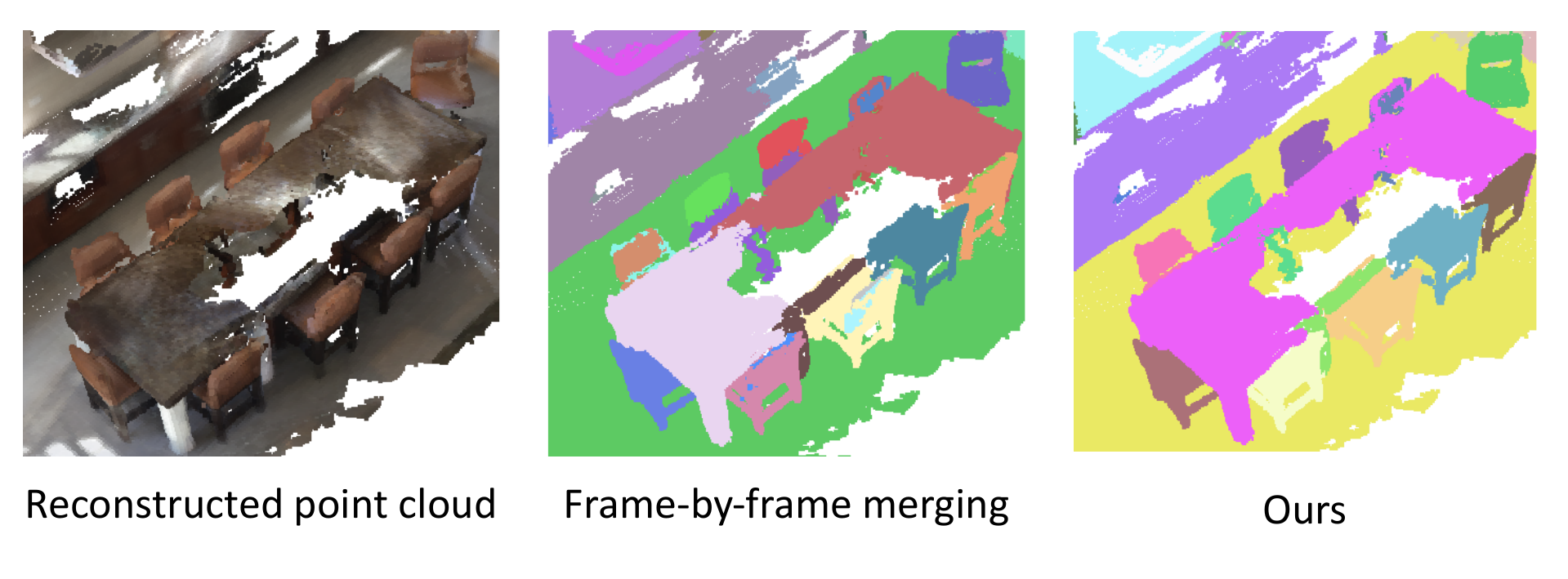}

   \caption{\textbf{Visual comparison of different merging strategies.} The "frame-by-frame" mask merging strategy struggles to handle segmentation for large instances and significant viewpoint disparity.}
   \label{fig:suppl_merge}
\end{figure}

%% file: table/table_suppl_merge.tex
\begin{table}[h]

\centering
\setlength{\tabcolsep}{5pt}  
\renewcommand{\arraystretch}{1.2}  

\begin{tabular}{c|ccc}
\specialrule{1.0pt}{0pt}{0pt}  
  & $AP$ & $AP_{50}$ & $AP_{25}$ \\
\specialrule{1.0pt}{0pt}{0pt}

Frame-by-frame  & 14.7 & 29.0 & 44.3 \\
Ours & \textbf{18.6} & \textbf{36.1} & \textbf{53.5} \\

\specialrule{1.0pt}{0pt}{0pt}  
\end{tabular}

\caption{Comparison of different mask merging strategies.}
\label{tab:suppl_merge}

\end{table}

%% file: fig/fig_suppl_mappingtable.tex
\begin{figure}[h!]
  \centering
   \includegraphics[width=1.0\linewidth]{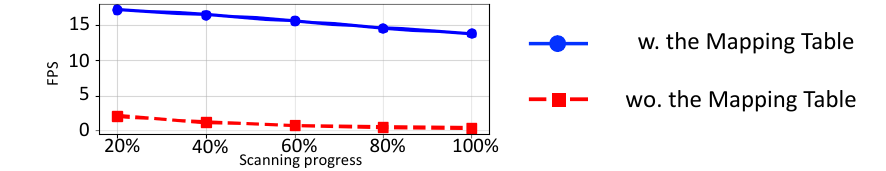}
   
   \caption{Comparison of speed with and without the mapping table.}
   \label{fig:suppl_mappingtable}
\end{figure}

%% file: sec/X_suppl_3_add_exp.tex
\section{Additional Experiments}
\label{sec:suppl_add_exp}

\subsection{More Full-sequence Results}
\input{table/table_suppl_sceneNN_per_seq}
The qualitative results and visual comparison of full-sequence instance segmentation are provided in the Sec. 4.2 and Sec. 4.3 of the main paper. To facilitate a more detailed comparison, we present the per-sequence results of our method and other online methods on SceneNN~\cite{hua2016scenenn}, as shown in \cref{tab:supp_per_seq}. 

SAM3D~\cite{yang2023sam3d} performs poorly overall due to its naive mask merging strategy. Compared with EmboidiedSAM~\cite{xu2024embodiedsam}, our method achieves higher performance under low and medium IoU thresholds but performs worse under high IoU thresholds. The reason is that as a learning-based method, EmbodiedSAM trains a classification head to to distinguish between foreground and background in advance. Background instances, such as walls and floors, are filtered out, leaving only foreground instances to be processed and merged. In contrast, as zero-shot methods, both our method and MaskClustering~\cite{yan2024maskclustering} lack pre-filtering operations and process all detected instances equally. This phenomenon is evident in Fig. 6 of the main paper, where the background is pained uniformly in gray. Since, under high IoU thresholds, the predicted walls and floors often fail to match the ground truth accurately, zero-shot methods may experience a drop in performance.

\subsection{More Intermediate Results}

In addition to the final segmentation results, we evaluate the intermediate segmentation outputs of EmbodiedSAM and our method, with results presented in Fig. 4 and Tab. 2 of the main paper. Furthermore, we conduct more detailed experiments to explore the core reasons behind the superior real-time segmentation performance of our method.

The real-time segmentation results at different quarters of the input sequence are presented in \cref{tab:supp_inter_result}. Each sub-table shows the AP scores at the completion of 25\%, 50\%, and 75\% of the sequence respectively. Our method outperforms EmbodiedSAM in most sequences, demonstrating the effectiveness of our global-consistent merging strategy. This also indicates that our approach has less reliance on post-processing steps, such as smoothing, to achieve high-quality segmentation results.

Meanwhile, we also provide a detailed visual comparison of instance segmentaion results on progressively scanned scenes in \cref{fig:supp_gallery_inter}. Several challenging scenes from ScanNet200~\cite{rozenberszki2022language} and SceneNN~\cite{hua2016scenenn} are showcased to highlight the performance of our method in complex environments. The intermediate segmentation results are directly presented on the reconstructed point cloud (EmbodiedSAM) or reconstructed mesh (ours), with backgrounds painted in gray. We observe that while EmbodiedSAM performs well on ScanNet200 (the first and fourth scenes), the dataset on which it is trained, it tends to output noisy segmentation results when transferred to other datasets (the other four scenes), as highlighted by the red bounding boxes in \cref{fig:supp_gallery_inter}.

The reason for this is that the mask merging strategy used by EmbodiedSAM is essentially a "frame-by-frame" approach. It utilizes the encoded features output by the trained model, along with the IoU of mask bounding boxes, to evaluate the similarity between newly detected masks and previous instance masks. This approach avoids calculating precise 3D spatial overlap between mask pairs, which may lead to issues when transferring to new datasets with different characteristics. In contrast, our mask merging strategy incorporates more global information during each merging step, fundamentally differing from the "frame-by-frame" approach, as discussed in \cref{subsec:supp_comp_fbf}.

\input{table/table_suppl_inter_bkup}


\subsection{More Ablation Studies}

We also test different $\tau_\text{weight}$ values and the results are shown in \cref{tab:suppl_mask_weight}, demonstrating the robustness of our method and confirming $\tau_\text{{weight}} = 5$ as the optimal parameter choice.

\input{table/table_suppl_weight}

%% file: table/table_suppl_sceneNN_per_seq.tex
\begin{table*}[h]

\centering
\setlength{\tabcolsep}{8.5pt}  
\renewcommand{\arraystretch}{1.2}  

\begin{tabular}{c|ccc|ccc|ccc}
\specialrule{1.0pt}{0pt}{0pt}  
\multirow{2}*{\textbf{Scene}} & \multicolumn{3}{c|}{\textbf{SAM3D$^*$}~\cite{yang2023sam3d}} & \multicolumn{3}{c|}{\textbf{EmbodiedSAM$^{**}$}~\cite{xu2024embodiedsam}} & \multicolumn{3}{c}{\textbf{Ours$^*$}} \\
~ & $AP$ & $AP_{50}$ & $AP_{25}$ & $AP$ & $AP_{50}$ & $AP_{25}$ & $AP$ & $AP_{50}$ & $AP_{25}$  \\

\specialrule{1.0pt}{0pt}{0pt}

005 & 9.1 & 25.6 & 54.8 & 15.5 & 25.9 & 48.4 & \textbf{21.0} & \textbf{49.5} & \textbf{63.4} \\
011 & 29.1 & 44.5 & 53.0 & \textbf{44.1} & 53.0 & 57.9 & 33.3 & \textbf{54.5} & \textbf{64.9} \\
015 & 6.4 & 16.9 & 32.7 & \textbf{17.8} & 32.3 & 43.3 & 16.9 & \textbf{35.7} & \textbf{45.1} \\
030 & 4.1 & 10.0 & 38.2 & \textbf{28.0} & \textbf{42.7} & \textbf{52.6} & 17.1 & 29.2 & 50.0 \\
054 & 9.3 & 27.0 & 50.7 & \textbf{23.5} & \textbf{41.4} & 61.4 & 16.9 & 31.7 & \textbf{61.2} \\
080 & 9.0 & 22.4 & 54.3 & 7.5 & 17.6 & 30.0 & \textbf{13.6} & \textbf{27.8} & \textbf{54.1} \\
089 & 5.9 & 17.2 & 48.2 & \textbf{9.7} & \textbf{22.3} & 42.3 & 7.8 & 16.1 & \textbf{50.4} \\
093 & 12.5 & 23.8 & 46.2 & \textbf{24.7} & 37.9 & 47.3 & 19.8 & \textbf{41.1} & \textbf{65.5} \\
096 & 9.9 & 22.5 & 55.8 & \textbf{21.7} & \textbf{36.2} & 44.5 & 16.4 & 29.2 & \textbf{60.6} \\
243 & 4.5 & 14.2 & 40.9 & \textbf{16.3} & 27.6 & 51.4 & 12.7 & \textbf{31.7} & \textbf{69.1} \\
263 & 14.8 & 39.2 & 59.0 & \textbf{27.9} & 38.9 & 47.3 & 26.6 & \textbf{52.6} & \textbf{66.7} \\
322 & 16.4 & 31.8 & 51.3 & \textbf{33.1} & 41.1 & 52.8 & 31.7 & \textbf{56.9} & \textbf{77.3} \\

\specialrule{0.25pt}{0pt}{0pt}

Overall & 9.1 & 21.3 & 43.4 & \textbf{20.1} & 32.5 & 46.3 & 18.1 & \textbf{35.3} & \textbf{59.5} \\

\specialrule{1.0pt}{0pt}{0pt}  
\end{tabular}

\vspace{5pt}
\caption{\textbf{Full-sequence instance segmentation results on SceneNN~\cite{hua2016scenenn}.} We present the per-sequence results of the online methods. $^*$: Zero-shot method, $^{**}$: Learning-based method.}

\label{tab:supp_per_seq}

\end{table*}

%% file: table/table_suppl_inter_bkup.tex
\begin{table}[ht]
    \centering
    \small
    \setlength{\tabcolsep}{4pt}  
    \renewcommand{\arraystretch}{1.1}  

    \begin{subtable}{0.45\textwidth}
        \centering
        \caption{Intermediate results at 25\% sequence completion.}
        \label{tab:example-a}
        \begin{tabular}{c|ccc|ccc}
            \specialrule{1.0pt}{0pt}{0pt}  
            \multirow{2}*{\textbf{Scene}} & \multicolumn{3}{c|}{\textbf{EmbodiedSAM$^{**}$}~\cite{xu2024embodiedsam}} & \multicolumn{3}{c}{\textbf{Ours$^*$}} \\
            ~ & $AP$ & $AP_{50}$ & $AP_{25}$ & $AP$ & $AP_{50}$ & $AP_{25}$  \\
            \specialrule{1.0pt}{0pt}{0pt}
            005 & 7.1 & 17.5 & \textbf{51.0} & \textbf{8.6} & \textbf{26.2} & 47.1 \\
            011 & 11.5 & 25.7 & 42.9 & \textbf{34.7} & \textbf{52.5} & \textbf{61.3} \\
            015 & 14.7 & 34.1 & 47.3 & \textbf{41.6} & \textbf{61.5} & \textbf{81.1} \\
            030 & \textbf{12.2} & 20.9 & 37.4 & 9.3 & \textbf{25.7} & \textbf{48.2} \\
            054 & \textbf{14.3} & \textbf{45.0} & 64.3 & 13.0 & 36.7 & \textbf{65.6} \\
            080 & 7.8 & 13.0 & \textbf{52.4} & \textbf{22.0} & \textbf{49.1} & 49.3 \\
            089 & \textbf{6.6} & \textbf{23.8} & 40.4 & 3.9 & 9.7 & \textbf{44.9} \\
            093 & 17.8 & 46.6 & 63.2 & \textbf{32.5} & \textbf{51.2} & \textbf{72.6} \\
            096 & \textbf{11.7} & \textbf{36.8} & 38.3 & 10.4 & 19.1 & \textbf{59.5} \\
            243 & 11.1 & 22.9 & \textbf{62.5} & \textbf{18.6} & \textbf{46.5} & 59.6 \\
            263 & 26.5 & 52.2 & 61.5 & \textbf{42.5} & \textbf{73.2} & \textbf{86.6} \\
            322 & 9.3 & 22.2 & 35.0 & \textbf{18.2} & \textbf{36.7} & \textbf{61.1} \\
            \specialrule{1.0pt}{0pt}{0pt}  
        \end{tabular}
    \end{subtable}

    \begin{subtable}{0.45\textwidth}
        \centering
        \caption{Intermediate results at 50\% sequence completion.}
        \label{tab:example-a}
        \begin{tabular}{c|ccc|ccc}
            \specialrule{1.0pt}{0pt}{0pt}  
            \multirow{2}*{\textbf{Scene}} & \multicolumn{3}{c|}{\textbf{EmbodiedSAM$^{**}$}~\cite{xu2024embodiedsam}} & \multicolumn{3}{c}{\textbf{Ours$^*$}} \\
            ~ & $AP$ & $AP_{50}$ & $AP_{25}$ & $AP$ & $AP_{50}$ & $AP_{25}$  \\
            \specialrule{1.0pt}{0pt}{0pt}
            005 & 8.5 & 17.4 & 42.3 & \textbf{12.9} & \textbf{34.8} & \textbf{60.2} \\
            011 & 15.8 & 35.1 & 55.2 & \textbf{32.0} & \textbf{61.2} & \textbf{68.1} \\
            015 & 17.2 & 37.9 & 38.1 & \textbf{20.4} & \textbf{41.1} & \textbf{52.3} \\
            030 & 14.3 & 22.7 & 40.1 & \textbf{17.1} & \textbf{30.0} & \textbf{48.5} \\
            054 & 19.7 & \textbf{60.6} & \textbf{76.3} & \textbf{20.4} & 43.0 & 69.4 \\
            080 & 5.1 & 16.5 & 35.6 & \textbf{17.8} & \textbf{40.6} & \textbf{57.2} \\
            089 & 4.8 & \textbf{18.9} & 41.0 & \textbf{6.1} & 16.9 & \textbf{49.0} \\
            093 & 16.7 & \textbf{39.9} & 59.7 & \textbf{21.0} & 38.6 & \textbf{63.2} \\
            096 & 10.0 & \textbf{35.3} & 42.8 & \textbf{15.3} & 26.4 & \textbf{61.1} \\
            243 & 9.1 & 20.2 & \textbf{63.8} & \textbf{10.9} & \textbf{31.9} & 60.8 \\
            263 & 29.4 & 52.2 & 69.2 & \textbf{38.6} & \textbf{68.3} & \textbf{76.7} \\
            322 & 11.2 & 24.7 & 37.5 & \textbf{16.9} & \textbf{42.9} & \textbf{74.1} \\
            \specialrule{1.0pt}{0pt}{0pt}  
        \end{tabular}
    \end{subtable}

    \begin{subtable}{0.45\textwidth}
        \centering
        \caption{Intermediate results at 75\% sequence completion.}
        \label{tab:example-a}
        \begin{tabular}{c|ccc|ccc}
            \specialrule{1.0pt}{0pt}{0pt}  
            \multirow{2}*{\textbf{Scene}} & \multicolumn{3}{c|}{\textbf{EmbodiedSAM$^{**}$}~\cite{xu2024embodiedsam}} & \multicolumn{3}{c}{\textbf{Ours$^*$}} \\
            ~ & $AP$ & $AP_{50}$ & $AP_{25}$ & $AP$ & $AP_{50}$ & $AP_{25}$  \\
            \specialrule{1.0pt}{0pt}{0pt}
            005 & 9.3 & 20.2 & 53.5 & \textbf{15.5} & \textbf{37.7} & \textbf{58.5} \\
            011 & 17.8 & 33.3 & 59.1 & \textbf{37.0} & \textbf{65.1} & \textbf{70.1} \\
            015 & 12.8 & 27.1 & 41.3 & \textbf{26.1} & \textbf{35.4} & \textbf{45.6} \\
            030 & \textbf{18.3} & \textbf{30.6} & \textbf{47.9} & 15.9 & 25.0 & 41.7 \\
            054 & 16.2 & \textbf{41.1} & \textbf{67.5} & \textbf{19.4} & 40.2 & 61.6 \\
            080 & 3.1 & 7.2 & \textbf{33.9} & 14.2 & \textbf{30.8} & \textbf{51.2} \\
            089 & 5.0 & \textbf{19.0} & 45.5 & \textbf{5.1} & 18.3 & \textbf{52.6} \\
            093 & 13.6 & 31.0 & 58.8 & \textbf{19.8} & \textbf{34.7} & \textbf{63.0} \\
            096 & 11.5 & \textbf{41.5} & 49.0 & \textbf{13.8} & 25.4 & \textbf{66.0} \\
            243 & 11.7 & 23.1 & \textbf{66.7} & \textbf{16.1} & \textbf{40.5} & \textbf{66.7} \\
            263 & 22.4 & 39.2 & 57.1 & \textbf{35.2} & \textbf{62.2} & \textbf{74.2} \\
            322 & 18.3 & 38.0 & 47.2 & \textbf{26.3} & \textbf{49.8} & \textbf{77.1} \\
            \specialrule{1.0pt}{0pt}{0pt}  
        \end{tabular}
    \end{subtable}

\caption{\textbf{Intermediate instance segmentation results on SceneNN~\cite{hua2016scenenn}.} The instance segmentation results are evaluated by mapping from the reconstructed point cloud or mesh to ground truth point cloud through point correspondences. $^*$: Zero-shot method, $^{**}$: Learning-based method.}
\label{tab:supp_inter_result}
\end{table}

%% file: table/table_suppl_weight.tex
\begin{table}[H]
\captionsetup{aboveskip=0pt, belowskip=0pt} 
\centering
\small

\begin{tabular}{l|cccccc}
\specialrule{1.0pt}{0pt}{0pt}  
$\tau_{\text{weight}}$  & 3 & 4 & 5 & 6 & 7 & 8 \\
\specialrule{0.5pt}{0pt}{0pt}


$AP_{25}$ & 52.1 & \textbf{53.5} & \textbf{53.5} & \textbf{53.5} & 53.0 & 52.6 \\
$AP_{50}$ & 32.7 & 36.1 & 36.1 & 36.1 & \textbf{36.3} & 34.1 \\
$AP$ & 17.9 & \textbf{18.6} & \textbf{18.6} & 18.3 & 18.0 & 17.4 \\

\specialrule{1.0pt}{0pt}{0pt}  
\end{tabular}

\vspace{5pt}
\caption{Ablation study on $\tau_{\text{weight}}$ on ScanNet200.}
\label{tab:suppl_mask_weight}

\end{table}

%% file: sec/X_suppl_4_video.tex
\section{Online Video Demo}
\label{sec:suppl_video}

\input{fig/fig_suppl_sensor}
We also provide a video demonstration, showcasing the real-time reconstruction and segmentation process of our method on several challenging scenes from ScanNet200~\cite{dai2017scannet, rozenberszki2022language} and SceneNN~\cite{hua2016scenenn}. In addition, we deploy our online segmentation method on an E05 robotic arm~\cite{E05}, guided by the Water II vehicle~\cite{water_two}, with a Microsoft Azure Kinect DK RGB-D sensor~\cite{azure_kinect} mounted at the end, as illustrated in \cref{fig:suppl_sensor}. A real-world demo is also presented in the video.

\input{fig/fig_supp_gallery_inter}

%% file: fig/fig_suppl_sensor.tex
\begin{figure}[h]
  \centering
  \includegraphics[width=0.85\linewidth, trim=0 0 0 0, clip]{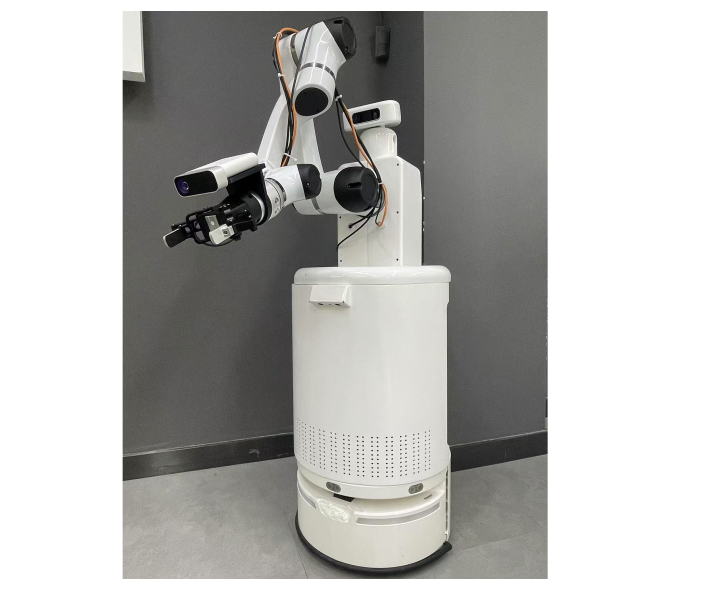}


   \vspace{10pt}
  \caption{\textbf{Equipment setup for real-world experiments.} The experimental setup consists of an E05 robotic arm~\cite{E05} guided by the Water II vehicle~\cite{water_two}, with a Microsoft Azure Kinect DK RGB-D sensor~\cite{azure_kinect} mounted at the arm's end.}
  \label{fig:suppl_sensor}
\end{figure}

%% file: fig/fig_supp_gallery_inter.tex


\begin{figure*}[hp] 
    \centering
    \begin{subfigure}[b]{\linewidth}
        \centering
        \includegraphics[width=0.95\linewidth]{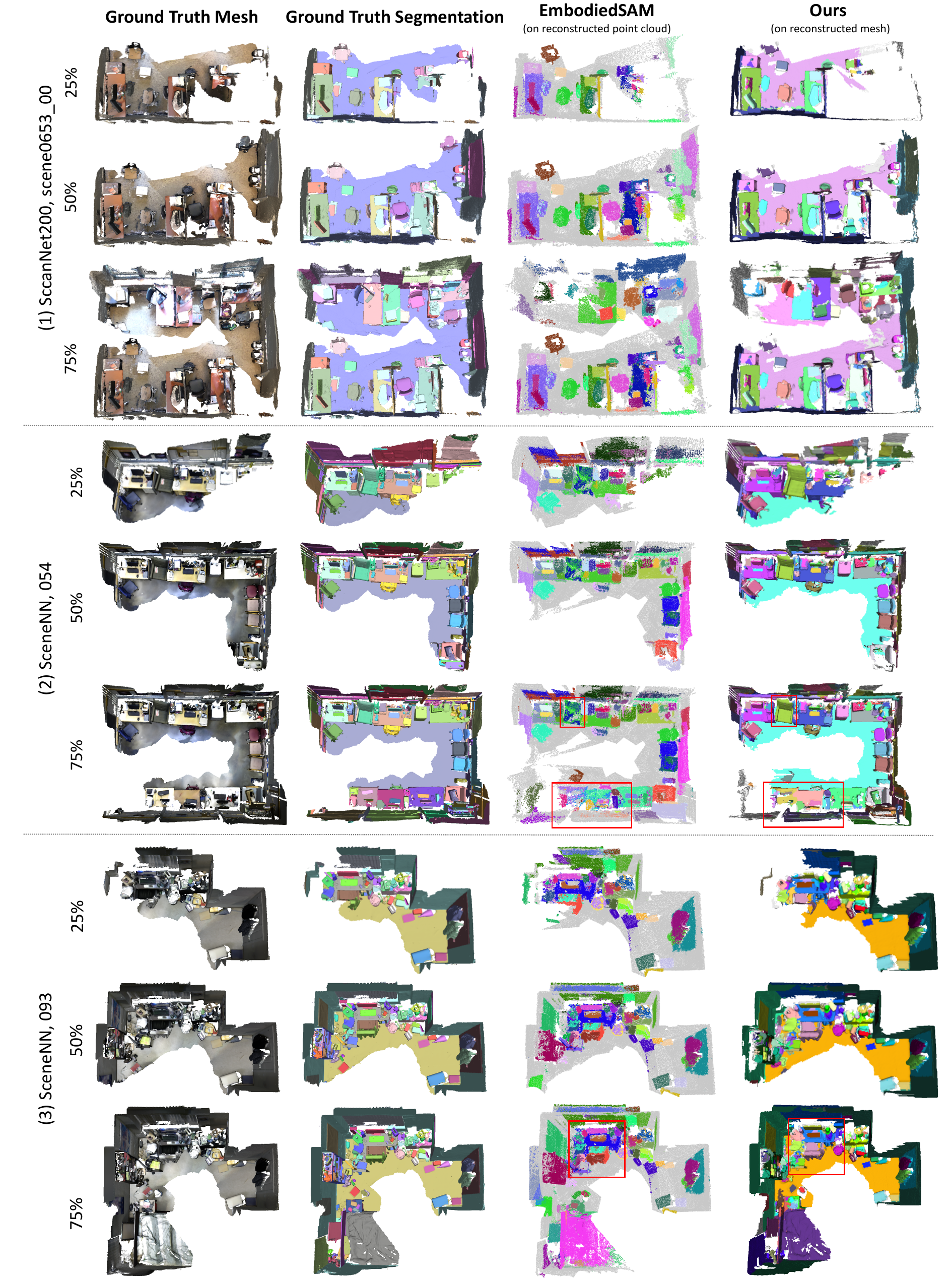}
        \label{fig:sub_a}
    \end{subfigure}
\end{figure*}

\begin{figure*}[hp] 
    \ContinuedFloat 
    \centering
    \begin{subfigure}[b]{0.975\linewidth}
        \centering
        \includegraphics[width=0.975\linewidth]{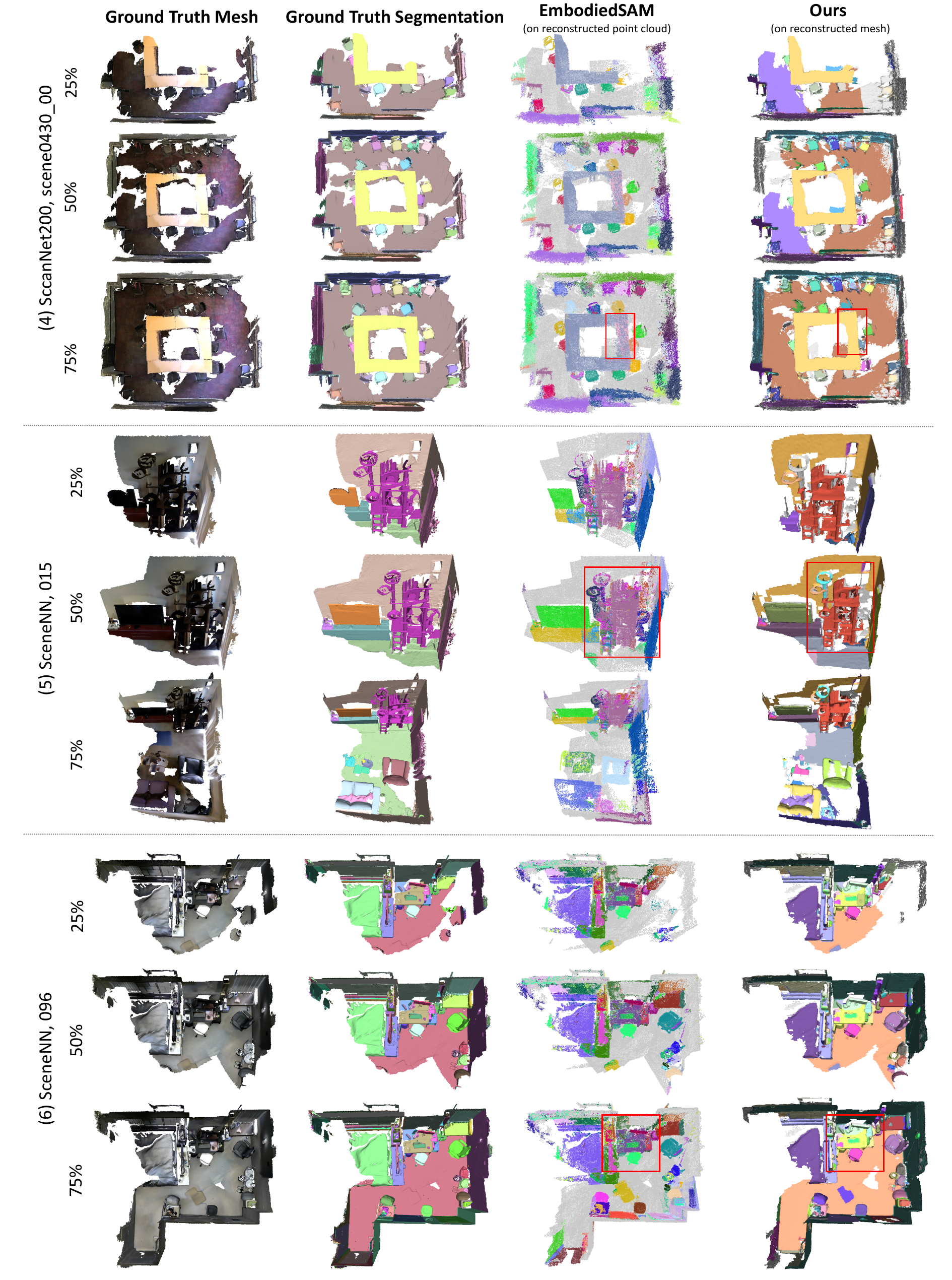}
        \label{fig:sub_b}
    \end{subfigure}
    
    \caption{Visual comparison of intermediate segmentation results with EmbodiedSAM~\cite{xu2024embodiedsam}.} 
    \label{fig:supp_gallery_inter}
\end{figure*}